\documentclass[10pt,twocolumn,letterpaper]{article}

\usepackage[pagenumbers]{cvpr} 

\usepackage[dvipsnames]{xcolor}

\usepackage{times}
\usepackage{epsfig}
\usepackage{graphicx}
\usepackage{amsmath}
\usepackage{amssymb}

\usepackage[utf8]{inputenc}
\usepackage[T1]{fontenc}
\usepackage{color,soul}
\usepackage{svg}
\usepackage{lipsum,graphicx,float}
\usepackage{wrapfig}
\usepackage{longtable}
\usepackage[mathcal]{euscript} 
\usepackage{tabularx}
\usepackage{booktabs}
\usepackage{enumitem}
\usepackage{comment}
\usepackage{siunitx}

\newcommand{\nSubjects}{15}
\newcommand{\ours}{Pulse3DFace}

\definecolor{cvprblue}{rgb}{0.21,0.49,0.74}
\usepackage[pagebackref,breaklinks,colorlinks,citecolor=cvprblue]{hyperref}

\def\paperID{229} 
\def\confName{3DV\xspace}
\def\confYear{2026\xspace}

\title{3D Blood Pulsation Maps}

\author{Maurice Rohr\\
	Technical University of Darmstadt\\
	{\tt\small rohr@kismed.tu-darmstadt.de} 
\and
Tobias Reinhardt\\
Technical University of Darmstadt\\
{\tt\small reinhardt@kismed.tu-darmstadt.de}
\and
Tizian Dege\\
Technical University of Darmstadt\\
{\tt\small dege@kismed.tu-darmstadt.de}
\and
Justus Thies\\
Technical University of Darmstadt\\
{\tt\small justus.thies@tu-darmstadt.de}
\and
Christoph Hoog Antink\\
Technical University of Darmstadt\\
{\tt\small hoogantink@kismed.tu-darmstadt.de}
}

\begin{document}

\twocolumn[{%
\renewcommand\twocolumn[1][]{#1}%
\maketitle
\vspace{-2em}
\centering
\includegraphics[clip, trim=0.7cm 0cm 0.55cm 0cm, width=\linewidth]{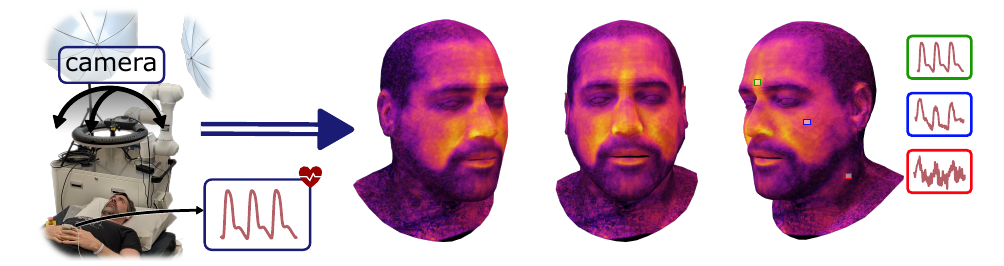}
\captionof{figure}{\ours\: is a dataset for estimating 3D blood pulsation maps. We showcase the dataset using the FLAME 3D-morphable model and various mappings for quasi-stationary pulse measurement. The maps can be used on avatars, animatable in terms of pose, expression, and view point as shown in the example renderings above, and to study dynamic skin perfusion. \vspace{1cm}}
\label{fig:teaser}
}]

\begin{abstract}
We present \ours, the first dataset of its kind for estimating 3D blood pulsation maps. 
These maps can be used to develop models of dynamic facial blood pulsation, enabling the creation of synthetic video data to improve and validate remote pulse estimation methods via photoplethysmography imaging. 
Additionally, the dataset facilitates research into novel multi-view-based approaches for mitigating illumination effects in blood pulsation analysis.
\ours\: consists of raw videos from 15 subjects recorded at 30\,Hz with an RGB camera from 23 viewpoints, blood pulse reference measurements, and facial 3D scans generated using monocular structure-from-motion techniques. 
It also includes processed 3D pulsation maps compatible with the texture space of the 3D head model FLAME. These maps provide signal-to-noise ratio, local pulse amplitude, phase information, and supplementary data.
We offer a comprehensive evaluation of the dataset’s illumination conditions, map consistency, and its ability to capture physiologically meaningful features in the facial and neck skin regions.
\end{abstract}    
\section{Introduction}\label{sec:intro}

Camera-based monitoring of the cardiovascular and respiratory systems has drawn much attention in recent years~\cite{McDuff.2023}.
First evidence that peripheral blood pulsation can be measured without contact using an imaging system was presented in 2000~\cite{Wu.2000}. 
In principle, photoplethysmography (PPG), used in smartwatches and pulsoximeters, relies on a controlled light source and measures the variations of light absorbed by the human skin due to the blood volume variations in superficial vessels during the cardiac cycle. Photoplethysmography Imaging~(PPGI)\footnote{Several names are used in the literature for this technique, among them remote PPG.} replaces the photodiode with a camera sensor and typically assumes ambient light instead of a controlled light source. This introduces additional complexity, as factors such as selection and tracking of an optimal region of interest (ROI) for pulse measurement, as well as various light-tissue interactions -- including scattering, reflection, and absorption -- must be considered.
PPGI might be able to replace contact sensors such as the PPG for estimating heart rate (HR) or respiration rate~\cite{McDuff.2023} in a wide range of areas, such as ICU-Monitoring~\cite{Wang.2024}, Arrhythmia-Detection~\cite{Li.2018}, and sports monitoring~\cite{Haugg.2022,Tan.2023}. 
Current research focuses on the utility of PPGI for heart rate variability (HRV)~\cite{Wang.2025}, blood pressure~\cite{Jeong.2016}, and (spatial) perfusion assessment~\cite{Lai.2022,Schraven.2023}, which is interesting because it enables the acquisition of entirely new information, such as in-surgery skin flap perfusion~\cite{Kossack.2022}, which cannot be captured with contact-based sensors. 
Under sufficiently illuminated low-motion conditions, the HR estimation task has many applicable solutions, irrespective of the approach, whether it is robust signal processing or deep learning~\cite{Scherpf.2024, Pstras.2025}.
However, the estimation of HR in real-world scenarios, as well as estimating HRV, blood pressure, or detecting arrhythmias is still a challenging task.

Due to data privacy regulations along with the inherent challenges of generating large-scale datasets, these problems remain difficult to address.
One solution is to augment existing data with underrepresented groups and difficult illumination settings~\cite{Ba.2022}, another one is to create entirely synthetic data of avatars showing \emph{plausible} blood volume pulse (BVP) to train end-to-end pulse signal extraction approaches, for example by varying the subsurface scattering color of a skin shader based on a reference pulse signal~\cite{McDuff.2022}. Despite their simplistic assumptions, both have shown promising results for HR estimation.
A different solution is to learn relevant facial regions on large image datasets, separately from the BVP extraction, to achieve the required robustness. 

Both the synthetic data generation pipeline and the ROI approach would benefit from a more principled data basis that realistically describes the distribution of pulsation across the facial skin. For example,~\cite{McDuff.2022} assume a constant phase of the BVP signal across the face and modulate its amplitude with mappings inspired by skin thickness values obtained from literature. Yet, the phase distribution can affect HRV measurement and shape the morphology of the BVP signal, which is crucial for detecting certain diseases~\cite{Bonnet.2022}.
%

%
Although studies have been conducted analyzing the optimal regions of the face to extract BVP from (see ~\Cref{sec:related_work}), we identified the following problems: (1) Most works employ non-uniform illumination or do not take the influence of illumination into account. This might lead only to the best illuminated (and not necessarily physiologically optimal) regions to be considered. (2) Measurements employ only a single point of view and thus cannot be translated to a 3D synthetics pipeline. (3) There do not exist quantitative metrics and datasets to assess the quality of the extracted pulse maps.


Hence, we propose the estimation of \emph{high resolution 3D blood pulsation maps} covering the whole face.
Our key contributions are:
\begin{enumerate}[noitemsep,topsep=0pt,parsep=0pt,partopsep=0pt]
    \item A pipeline for creating 3D skin perfusion models building on FLAME~\cite{Li.2017}, an easily animatable model compatible with most available 3D Software. We see its main purpose in the creation of artificial data, improving on~\cite{McDuff.2022}, e.g. for training end-to-end BVP extraction approaches.
    \item A multi-view pulsation dataset that mitigates illumination biases present in previous works by capturing videos from multiple angles. We cover phase, amplitude, and signal-to-noise ratio~(SNR) computed using robust algorithms, and enable the quantification of illumination effects on BVP extraction, for which we propose a first measure.
\end{enumerate}

\noindent
We make the complete dataset publicly available, offering researchers a valuable resource for benchmarking, method development, and reproducible experiments.
It is available after registration\footnote{Dataset-Website: \url{https://github.com/KISMED-TUDa/pulse3dface}} in accordance with GDPR rules to protect the participant's privacy. 
\section{Related Work}
\label{sec:related_work}

Although we are the first to perform 3D pulse map reconstruction, there is extensive, individual literature on perfusion mapping, related datasets, and 3D face reconstruction.

\paragraph{Perfusion Mapping:}
We focus on camera-based techniques for perfusion mapping and refer to~\cite{Hammer.2022} for a comparison to alternatives.
Traditionally, perfusion mapping shares common methods with region-of-interest (ROI) selection for PPGI, which is why we show recent works for both here.
One of the first perfusion mapping studies using cameras calculates a phase-shifted correlation matrix for each video pixel with a reference signal to construct a dynamic pulse map~\cite{Kamshilin.2011}. 
Most works build pulse maps from a sliding window by computing quasi-stationary values from the pulse signal such as the signal-to-noise ratio (SNR)~\cite{Haan.2013}, pulse amplitude and phase~\cite{Moco.2017,Teplov.2014,Zaunseder.2018}. 
Typically, the measurement setup consists of a global shutter camera that records uncompressed video at 20--100\,FPS~\cite{Kamshilin.2011,Hammer.2022} and an illumination source that is based on LEDs~\cite{Kossack.2022,Hammer.2022} or high-frequency fluorescent lamps~\cite{Moco.2017,Moco.2018}. 
The cameras are either monochrome in combination with narrow spectrum green light~\cite{Kamshilin.2011,Teplov.2014} or RGB in combination with white light~\cite{Moco.2017,Kossack.2022,Lai.2022,Schraven.2023}. 
Earlier studies use a cross-polarization scheme to reduce specular reflections~\cite{Teplov.2014,Moco.2018}.
Usually, the illumination is parallel to the camera axis~\cite{Kamshilin.2011,Kossack.2022,Hammer.2022} to optimally capture the pulsatile absorption characteristic of the skin resulting in the BVP signal. 
However, the desired signal source can also be mechanical, such as the cardiac related skin motion at large arteries, in which case the main illumination is orthogonal to the camera axis~\cite{Moco.2017,Moco.2018}. 
This \emph{mechanical signal} is best extracted from the red color channel of an RGB camera~\cite{Moco.2018}.
The BVP is estimated either by using a single channel that covers absorption optimally, e.g. green~\cite{Kamshilin.2011,Zaunseder.2018,Hammer.2022}
or a robust method that reduces specular reflection~\cite{Kossack.2022,Lai.2022,Hammer.2022} such as the Plane-orthogonal-to-skin algorithm (POS)~\cite{Wang.2017}. 
The resulting signal is then compared to a \emph{reference signal} using correlation~\cite{Frassineti.2017}, typically preceded by a Hilbert transform of the signal~\cite{Moco.2017,Kossack.2022,Lai.2022}. 
Hammer et al.~\cite{Hammer.2022} forgo the correlation and compute the amplitude of the signal from peak-through differences.
Otherwise, the signal is band-limited to the typical HR range of the application scenario, e.g. \qty{0.6}{\hertz} to \qty{4.0}{\hertz}~\cite{Kossack.2022}.
The reference signal can be a contact PPG~\cite{Moco.2017}, a POS signal extracted from a reference region~\cite{Kossack.2022,Lai.2022}, or a sine wave oscillating with the reference HR~\cite{Zaunseder.2018}.
Summary parameters computed from pulse maps such as the arterial stiffness index, wave reflection parameters~\cite{Moco.2017}, as well as the global perfusion index~\cite{Hammer.2022,Kossack.2022,Schraven.2023,Lai.2022} have been used to assess the local tissue, for example in intraoperative fasciocutaneos flap monitoring~\cite{Schraven.2023,Kossack.2022}.
Furthermore, phase maps enable the differentiation of proximal and distal ends of the intestine during surgery~\cite{Lai.2022} and can be use to estimate a blood flow field~\cite{Yang.2015} when removing motion components. 
Pulse transit time (PTT), which shows correlation with blood pressure~\cite{Jeong.2016} was measured between distal locations (hand and face), by using amplitude maps to select suitable regions for PPGI computation~\cite{Shao.2014}.

Perfusion mapping can be used to select ROIs for robust BVP signal extraction~\cite{Rohr.2024} and was first used by~\cite{Lempe.2013} to identify the cheek and forehead as the regions with the highest signal amplitude. 
\cite{Kim.2021} proposed a BVP similarity metric between manually defined super-regions of the facial skin and a reference PPG. 
They found equal influence of both the number of pixels of a region and the region's average skin thickness reference values for pulse estimation performance.

\emph{Static perfusion models}, different to the dynamic mappings, focus on superficial perfusion of the face caused by dilation or constriction of blood vessels. 
Usually, they are used to induce noticeable changes of skin color for facial animations.
This can be achieved by obtaining facial maps of hemoglobin concentration during different emotions~\cite{Jimenez.2010} and integrating it in a setup for facial appearance model acquisition~\cite{Gotardo.2018}. 
Another application is cost-efficient spatial blood-oxygenation and blood volume fraction measurement~\cite{Nogue.2024}.

Except for static perfusion models, most of the dynamic/stationary perfusion measurements do not take the 3D surface of the face and different viewing angles or illumination effects into consideration.

\begin{table}[t]
    \setlength{\tabcolsep}{2.5pt}
    \centering
    
    \resizebox{\columnwidth}{!}{\begin{tabular}{lcccccccc}
        \toprule
        Dataset & \#Subj. & Public  & Resolution & Fps & Motion & Comp. & Muli View & Ref.\\
        \midrule
        
        BP4D+~\cite{Zhang.2016} & 140 & \checkmark & 1040 x 1392 & 25 & X & X & (\checkmark) & BP \\
        OBF~\cite{Li.2018} & 106 &  & 1920 x 1080 & 60 &  & X &  & BVP \\
        MR-NIRP~\cite{Nowara.2018} & 12 & \checkmark & 640 x 640 & 30 &  &  &  & BVP \\
        UBFC-Phys~\cite{Sabour.2023} & 56 & \checkmark & 1024 x 1024 & 35 &  & X &  & BVP\\
        NERsemble\cite{Kirschstein.2023} & 222 & \checkmark & 3208 x 2200 & 73 & X & X & \checkmark & - \\
        \midrule 
        Ours & \nSubjects & \checkmark & 1224 x 1024 & 30 &  &  & \checkmark & BVP \\
        \bottomrule
    \end{tabular}}
    \caption{Existing video datasets of human faces suitable for pulse map computation. Note that for each dataset, we only count the best illuminated scenario with minimal movement. References can be blood pressure (BP) or blood volume pulse (BVP)}
    \label{tab:existing_datasets}
\end{table}


\begin{figure*}[ht!]
    \centering
    \includegraphics[width=0.9\linewidth]{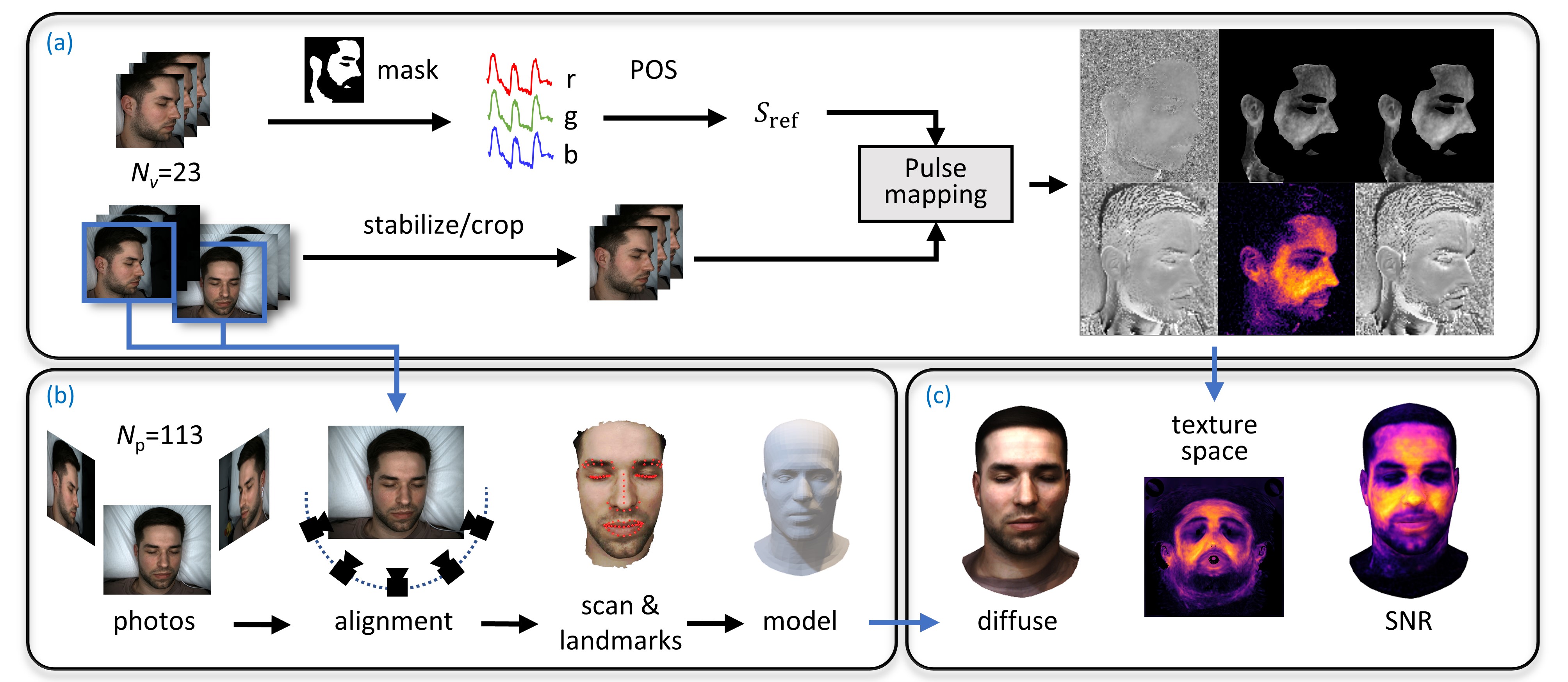}
    \caption{Overview of the computation of 3D pulse maps. (a) 2D pulsation map computation from videos. (b) Alignment of photos and first frame of videos via SFM and FLAME registration with the scan with landmarks followed by fitting. (c) Texture space computation. POS: plane-orthogonal-to-skin algorithm.
    }
    \label{fig:methods_figure}
\end{figure*}

\paragraph{Datasets:}
Several datasets provide facial videos, each covering some essential aspects for detailed pulsation mapping, see \Cref{tab:existing_datasets}.
The BP4D+~\cite{Zhang.2016} database contains 140 subjects who were filmed for one minute during 10 different emotional states. The setup consists of a 3D dynamic imaging system, a thermal camera, and studio illumination. However, the RGB video resulting from the 3D system is heavily compressed and contains movement due to the emotions.  
OBF~\cite{Li.2018} provides data of 100 healthy subjects and six atrial fibrillation patients in 5-minute recordings during resting and post-exercise state. It contains low-motion scenarios and features strong LED illumination. The dataset is temporarily not available.
MR-NIRP~\cite{Nowara.2018} and UBFC-Phys~\cite{Sabour.2023} contain raw or low-compression video with stable light conditions, and UBFC-Phys has been used for exploring ROI selection before.
UBFC-Phys~\cite{Sabour.2023} contains a rest task with 3 minute video of 56 healthy subjects illuminated by a uniform light source at 1\,m distance. The RGB stream is compressed with MJPEG and the database has been used before for exploring optimal ROI selection.
The NERsemble~\cite{Kirschstein.2023} dataset provides 222 subjects and was recorded with 16 cameras with high resolution and high frame rate. Due to its focus on 3D reconstructions of human faces in different poses and emotion states, the recordings are very short (ca. 8--15 seconds), compressed, contain heavy motion, and the dataset provides no BVP reference. 

To conclude, existing datasets are not suitable for 3D pulse map reconstruction. 

\paragraph{3D Reconstruction:} 
Facial 3D reconstruction approaches can be categorized into high resolution personalized face models and generic face models.
Personalized face models such as \cite{Alexander.2009} usually employ high resolution facial scans with a multi-view stereo camera setup and active illumination patterns that are used as a baseline by 3D artists to create a mesh optimized for animation. 
One of the first generic face models, also called 3D morphable models (3DMM), builds on the principal component analysis using head scans of 200 young adults~\cite{Blanz.1999}.
FLAME~\cite{Li.2017} additionally models facial expression and eyeballs explicitly, allows animation, and was trained on over 10.000 registered heads. It has become a backbone of many later models. 
DECA~\cite{Feng.2021} augments FLAME with an appearance model while also capturing wrinkles and pores (facial details) as part of the geometry and is able to reconstruct the head shape and appearance from single images. 
More recently, FLAME was used as an underlying model for analysis-by-synthesis approaches that employ deep learning~\cite{Zhang.2023,Retsinas.2024} and Gaussian splats~\cite{Qian.2024}, achieving remarkable detail and more realism of animation by employing physically motivated priors~\cite{Yang.2024}.
The reader is referred to~\cite{Zollhofer.2018} for an introduction on face reconstruction. 

In our study, we employ FLAME because it provides a single shape space for all 3D reconstructed faces, enabling the computation of statistics on the pulse maps (see supplemental). 

\section{Method}

Our proposed method \ours\: estimates high resolution 3D pulsation maps by employing statistical shape models created from RGB images and RGB video data of a subject. 
Specifically, we assume a dataset for each subject consisting of $n_p$ photographs covering the complete area of the face and $n_v$ videos taken from various angles (see \Cref{sec:dat_gen}).
The photographs are used to recover a detailed mesh, while the videos are employed to estimate the 2D pulsation maps that are lifted into 3D using the mesh.
The proposed procedure for estimating 3D pulsation maps is depicted in \Cref{fig:methods_figure} and detailed in the following.

\subsection{2D Pulsation Maps}
\label{sec:pulse_maps}

To estimate pulsation maps, we assume a quasi-stationary BVP signal with a locally constant amplitude and constant phase difference to a reference signal. The camera is assumed to consist of pixel sensors, where each sensor measures the same area of the skin for the duration of the estimation. We compute eight spatially resolved/pixel-wise parameters to generate the pulsation maps: SNR ($M_\text{SNR}$), phase of PPGI (POS, $M_{\text{p}}$), phase of raw color signals ($M_{\text{p},c}$) and color signal amplitude ($M_{\text{a},c}$) for each color channel $c \in [r,g,b]$, as well as the HR ($M_\text{HR}$) for validation.

\paragraph{Reference signal:}
We compute a reference signal $S_\text{ref}$ from the whole face instead of relying on the contact PPG to minimize location-specific morphology differences~\cite{Braun.2024}. Using deeplab-based skin segmentation~\cite{Scherpf.2021}, we average the skin pixels to an RGB signal and apply the POS~\cite{Wang.2017} method, resulting in $S_\text{ref}$.
The reference heart rate $\text{HR}_\text{ref}$ is defined as the frequency at the maximal power of the power spectrum of $S_\text{ref}$ and validated using the contact PPG.

\paragraph{Computation of pulsation maps:}
We compute all parameters in a $k \times k$ spatially sliding window with $k \in [3,5,7,9,13,17]$ and for seven overlapping \qty{20}{\second} segments $i$ extracted from the \qty{70}{\second} video to generate maps $M_{*}^i$, by averaging the window to a $[600,3]$ RGB-signal. Finally, $M_{*}^i$ are averaged to reduce the noise present in single parts of the recording.

SNR, phase, and HR are computed from the RGB signal by first applying POS to extract the BVP signal $S$. The HR is computed from this RGB signal identically to $\text{HR}_\text{ref}$. The phase is extracted from the index of the Fourier transformed (FFT) signal corresponding to HR\textsubscript{ref}.
The SNR is adapted from \cite{Haan.2013} and computed as:

\begin{equation}
\small
    \text{SNR} =  10\log_{10}\frac{\sum_{f \in \mathcal{F}_{\text{HR}}}(U_t(f)\hat{S}(f))^2}{\sum_{f \in \mathcal{F}_{\text{HR}}}(1-U_t(f))\hat{S}(f))^2} ~~,
\end{equation}

where $\mathcal{F}_{\text{HR}}$ captures the full physiological range of the HR between 30-200\,BPM in addition to its first harmonic and a certain variability. $\hat{S}$ is the Fourier transformed signal, and $U_t(f)$ is a window function around $\text{HR}_\text{ref}$ and its first harmonic. It is constructed to capture variations around the average HR:
\begin{equation}
\small
   U_t(f) = \begin{cases}
     1 & \text{if} \: |f-\text{HR}_\text{ref}|\leq6 \: \And \: | f-2\cdot\text{HR}_\text{ref}|\leq12 \\
     0 & \text{else}
   \end{cases}
\end{equation}

The maps $M_\text{p}$ and $M_\text{HR}$ are only used to judge the validity of the pulse signal estimates, as HR cannot vary spatially and the phase map should be similar to the color channel-based phase maps.

To compute phase and amplitude of each color channel $c$, we detrend the signal $S_c$ by applying a lowpass filter with cutoff frequency of 0.4\,Hz to result in $S_{c,\text{lp}}$ and compute
\begin{equation}
    \hat{S}_c=\frac{S_c-S_{c,\text{lp}}}{S_{c,\text{lp}}}.
\end{equation}
We apply a bandpass filter to $S_\text{ref}$ with filter frequencies (0.4--4\,Hz) proposed in~\cite{Moco.2017}. 
Based on~\cite{Moco.2017}, we compute the Hilbert transform of $S_\text{ref}$ and scale $\bar{S}_\text{ref,hilb} \propto S_\text{ref,hilb}$ such that
\begin{equation}
    \sum\Re\{\bar{S}_\text{ref,hilb}\}\bar{S}_\text{ref,hilb}=1.
\end{equation}
The amplitude $A_c$ and phase $P_c$ of the signal are
\begin{equation}
    P_c = \angle(\bar{S}_\text{ref,hilb}\hat{S}_c), ~~ A_c = |(\bar{S}_\text{ref,hilb}\hat{S}_c)|.
\end{equation}

\begin{figure}
    \centering
    \includegraphics[width=\linewidth]{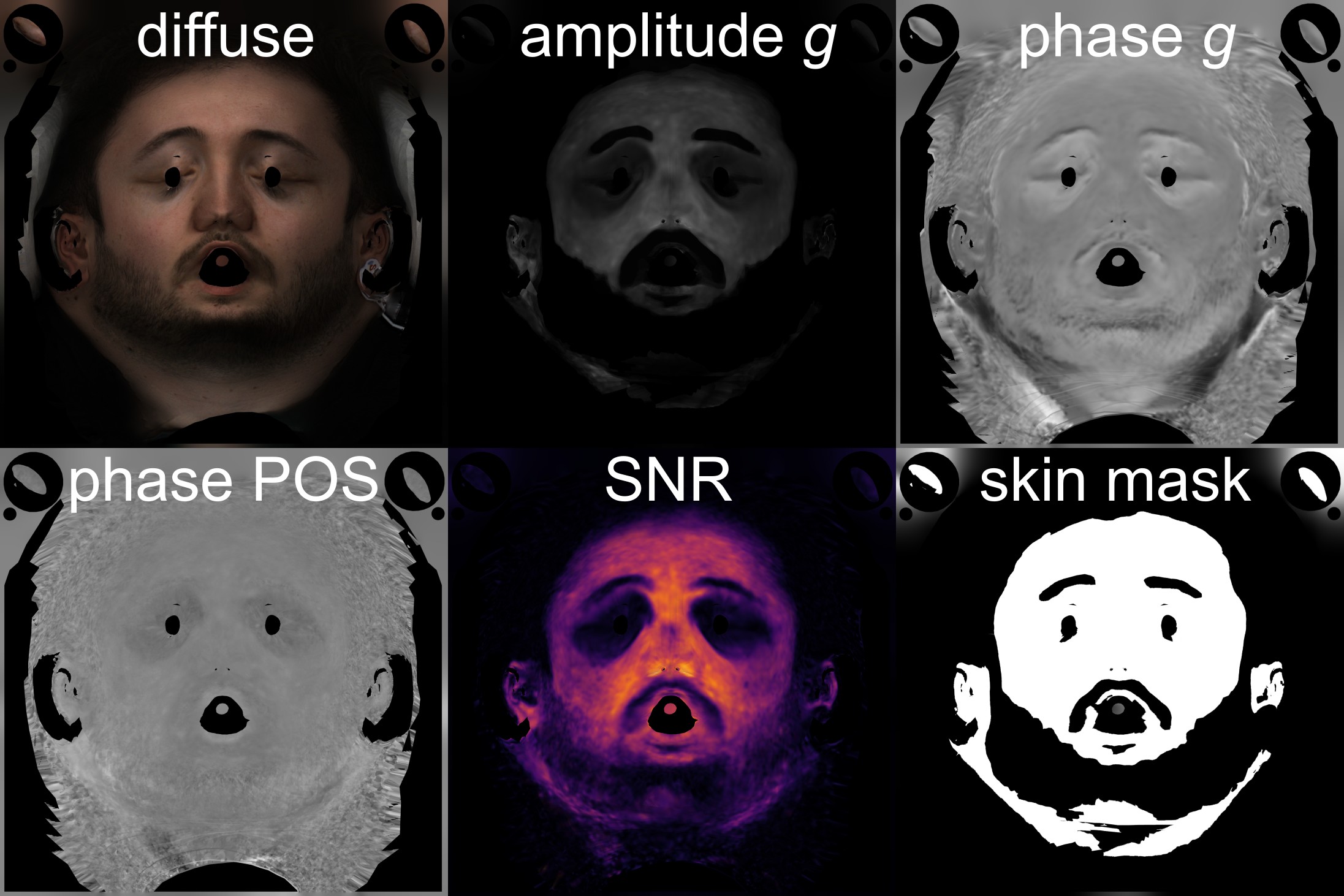}
    \caption{Reconstructed signals in FLAME texture space: Amplitude g and phase g refer to the green channel signal that shows the highest SNR compared to red and blue. The amplitude is shown with applied skin mask, as motion at facial boundaries and hair heavily influences the amplitude estimates.}
    \label{fig:texture_summary}
\end{figure}

\subsection{3D Pulsation Mapping}
Based on the captured $n_p$ photographs and $n_v$ videos, we use structure-from-motion (SFM)\footnote{Metashape, https://www.agisoft.com/} to obtain the corresponding $n_p+n_v$ camera poses and intrinsics as well as a textured surface mesh $\mathcal{T}$.
We clean and center this mesh to simplify further processing, and fit FLAME~\cite{Li.2017} to scan $\mathcal{T}$.
FLAME is a statistical shape model of the head with expressions, parameterized by the shape $\vec{\beta}$, pose $\vec{\theta}$, and expression $\vec{\psi}$ parameters, defined as $M(\vec{\beta},\vec{\theta},\vec{\psi})$.
For the initialization of the rigid transformation aligning both the FLAME mesh $\mathcal{M}$ and scan $\mathcal{T}$, we use MediaPipe's facial landmark model Blazeface~\cite{Bazarevsky.2019}.
We use pytorch3d\footnote{https://pytorch3d.org/} to render $\mathcal{T}$ to a 2D image, detect the landmarks, and backproject the landmarks onto $\mathcal{T}$, yielding 3D landmarks of $\mathcal{T}$. The corresponding landmarks on $\mathcal{M}$ are taken from the definition of~\cite{Zielonka.2022}.
Fitting $\mathcal{T}$ to $\mathcal{M}$ is done in three steps. First, $\mathcal{T}$ is scaled to match the size of $\mathcal{M}$. Second, the rigid transform $R,T$ from $\mathcal{M}$ to $\mathcal{T}$ is estimated based on the landmark positions and applied.
Third, the non-rigid fitting is performed using conjugate gradient iteration with the Chumpy~\cite{Loper.2014} implementation of the dogleg algorithm~\cite{Nocedal.2006} based on the public code base\footnote{https://github.com/Rubikplayer/flame-fitting}. 
During non-rigid fitting $T$, $\vec{\beta}$, $\vec{\theta}$, and $\vec{\psi}$ are optimized, using the following weights: vertex distance weight $\lambda_D=2.5$, landmark distance weight $\lambda_L=0.01$, shape regularizer $\lambda_\beta=0.0001$, expression regularizer $\lambda_\psi=0.0002$, pose regularizer $\lambda_\theta=0.001$, sigma value of Geman-McClure robustifier $\sigma_\text{GMO}$=0.0001. $R$ is fixed during that optimization. 

We use the optimized camera positions and the FLAME parameters to create a diffuse texture for $\mathcal{M}$. The result of this step is the textured FLAME model of the subject.
Analogously, we map the pulse maps computed from videos (as detailed in Section~\ref{sec:pulse_maps}) into the texture space of FLAME, see~\Cref{fig:texture_summary}. Spatial averaging is used to combine pulsation maps from different camera angles.
\begin{figure*}[htp]
    \centering
    \includegraphics[width=0.9\linewidth]{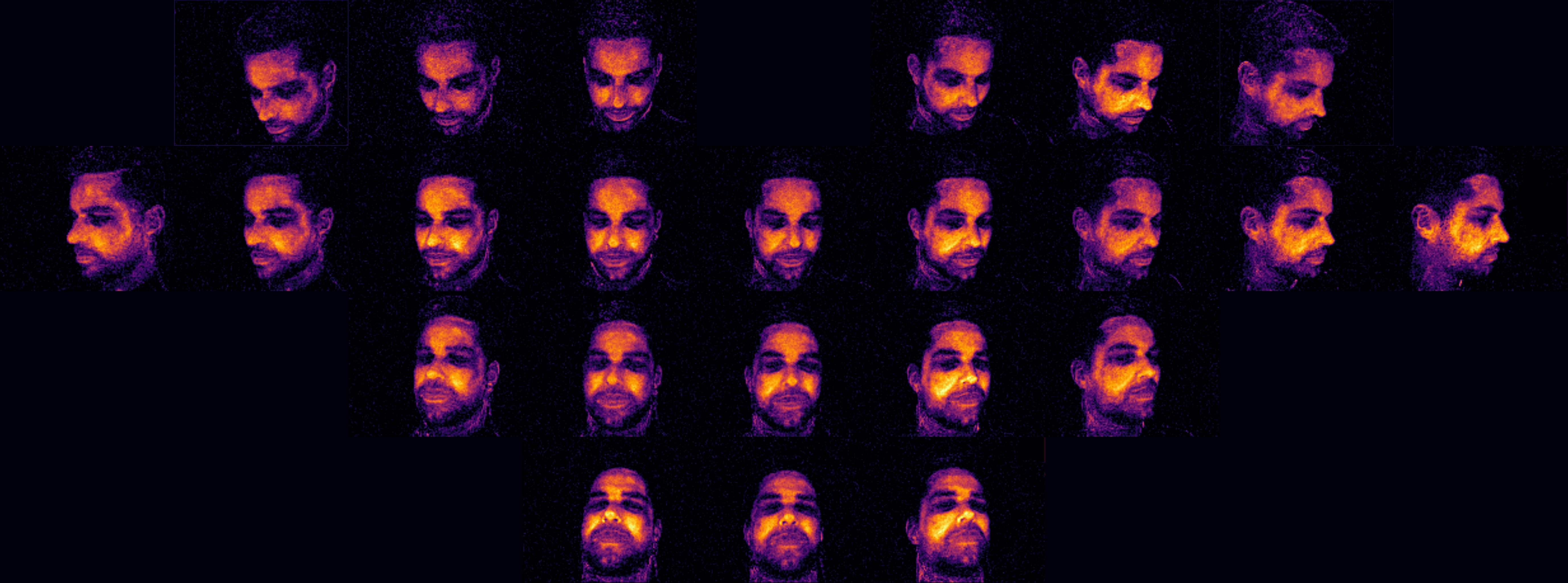}
    \caption{SNR maps from all 23 viewing directions for subject S01.}
    \label{fig:all_views_snr}
\end{figure*}

\section{Data Capture}
\label{sec:dat_gen}

\begin{figure}[b]
    \centering
    \includegraphics[width=\linewidth]{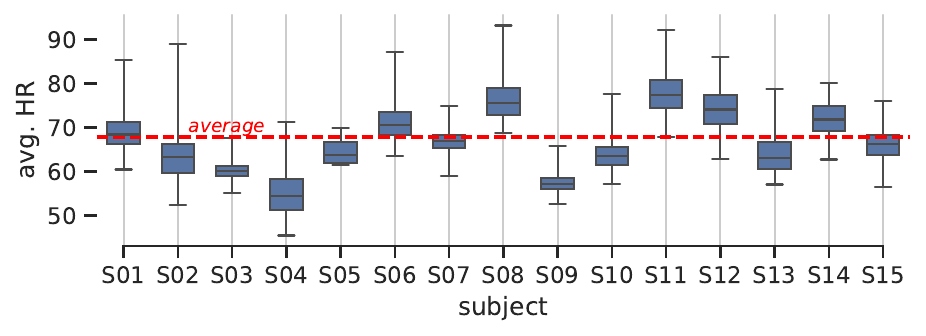}
    \caption{Heart rate averaged over 20\,s per subject in our dataset.}
    \label{fig:dataset_stats}
\end{figure}

\begin{figure*}[htp]
    \centering
    \begin{subfigure}[t]{0.45\textwidth}
        \centering
        \includegraphics[height=5.3cm]{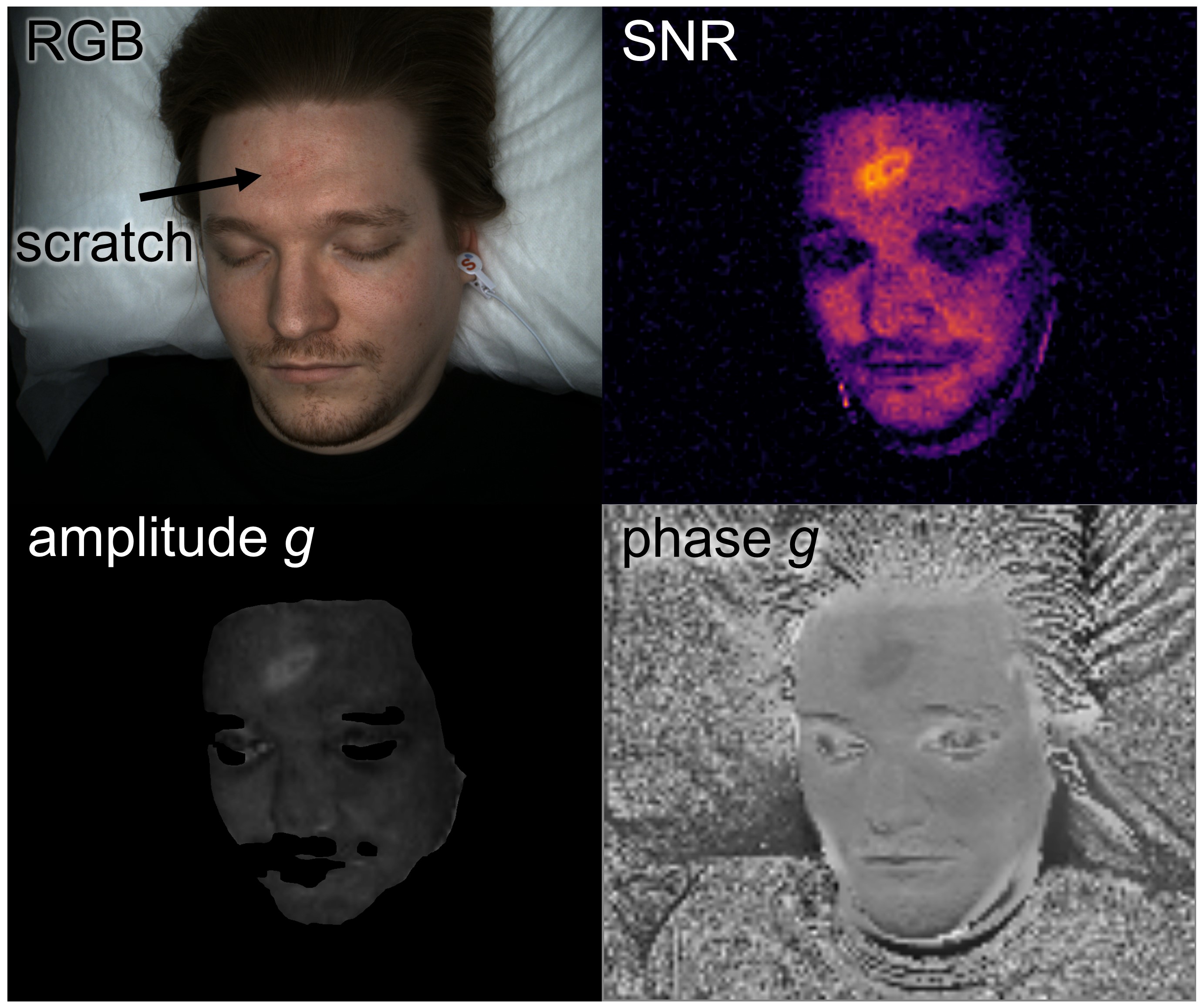}
        \caption*{(a)}
    \end{subfigure}
    \begin{subfigure}[t]{0.45\textwidth}
        \centering
        \includegraphics[height=5.3cm]{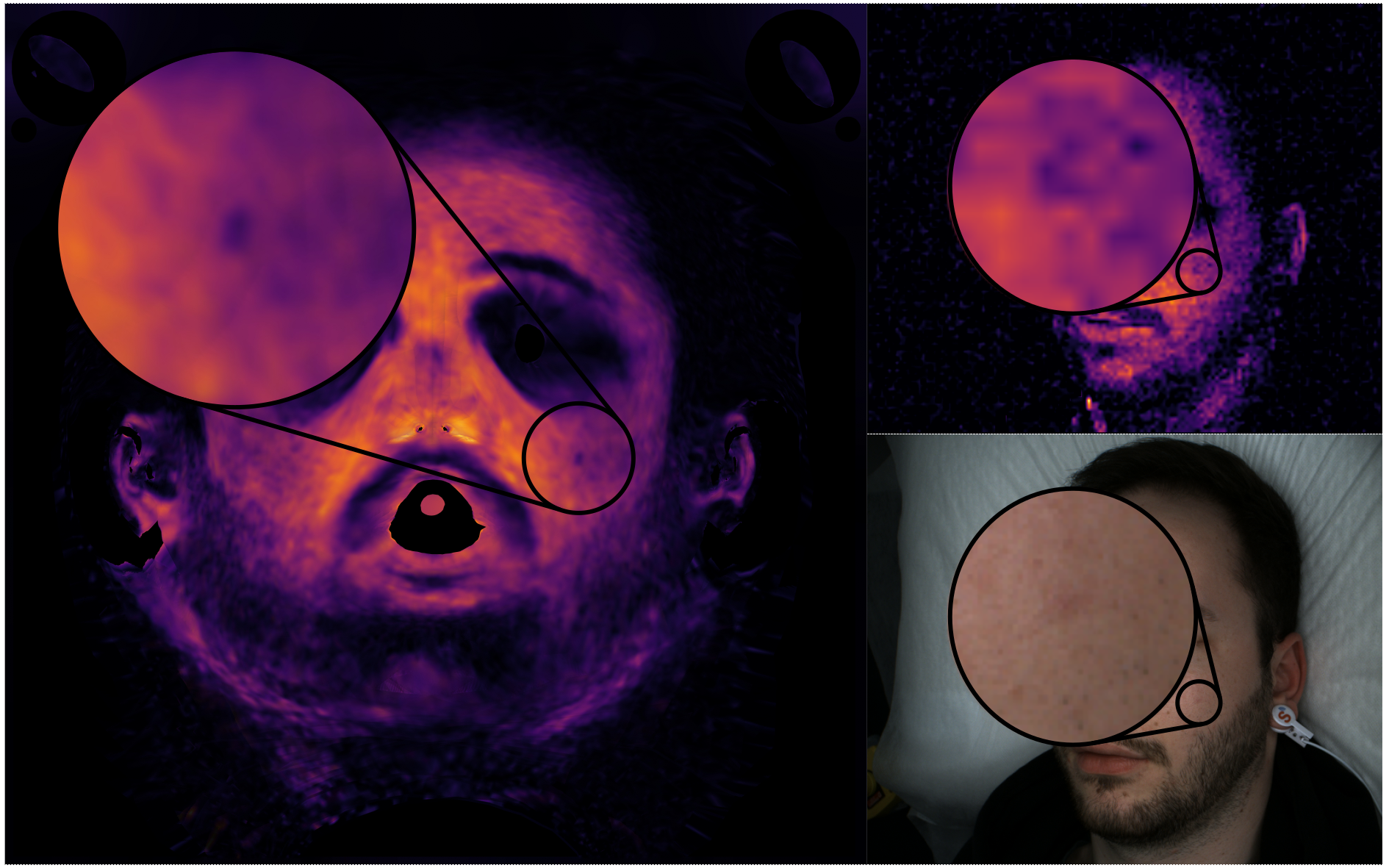}
        \caption*{(b)}
    \end{subfigure}
    \caption{Qualitative analysis. A slight skin scratch (a) increasing perfusion shows that pulse maps correspond to actual changes in superficial skin perfusion. Low perfusion regions, such as pimples (b), can be emphasized based on the 3D SNR map.}
    \label{fig:ex_validity}
\end{figure*}

Our dataset consists of \nSubjects\: adult healthy subjects (6 female, age 21-39 years) with no known heart conditions. Participants gave informed consent and the study was approved by the TU Darmstadt ethics committee (EK 01/2025). 
Due to the sensitivity of the captured data, all participants in our dataset signed an agreement form compliant with GDPR requirements.
Please note that GDPR compliance includes the right for every participant to request the timely deletion of their data. We will enforce these rights in the distribution of our dataset.

\paragraph{Optical recording setup:}
Our proposed method is based on the data captured with an optical sensor.
Specifically, we use a \textit{FLIR Blackfly RGB camera} (BFS-U3-50S5C-C) to record 30\,FPS video with a resolution of 1224 x 1024 pixel.
%
The focus and aperture are manually adjusted once for the measurement distance of 42\,cm and to assure a large depth of field as well as sufficient SNR for PPGI estimation. Accordingly, the exposure time is fixed to 3.5\,ms.
For illumination, we use two studio lights with softboxes (\textit{Proxistar}, LED 35\,W, 5500°K, softbox diameter 84\,cm) and a LED ring light (\textit{Walimex Pro Medow 960}, 96\,W, adjusted to 5500°K, outer diameter 45\,cm).
The camera and the LED ring light are attached to a robot arm (\textit{Doosan A0509 cobot}) which is used to change the camera poses (see ~\Cref{fig:teaser}).

\paragraph{Reference heart rate sensors:}
As a reference we use a \textit{Shimmer optical pulse ear-clip} and a \textit{biosignalsplux BVP} finger sensor with a sampling rate of 128\,Hz.
The reference sensors are synchronized with the video using UNIX-timestamps. 
The average HR changes over time and between subjects by 42\,BPM (see~\Cref{fig:dataset_stats}).

\paragraph{Recording procedure:}
The subject is lying in supine position on a hospital bed and is instructed to minimize head motion. 
First, a scan of 113 equidistantly spread images (step size 6°) is taken using a robot arm trajectory on a hemisphere around the head of the subject.
An approximate diffuse illumination setting is achieved by using only the soft boxes. Then, 70\,s videos are taken at 23 positions (step size 15°) with added ring light to provide strong illumination for the PPGI estimation. 
To visualize the captured data, \Cref{fig:all_views_snr} illustrates all 23 views with the SNR map.


\section{Results}

Our systematic analysis of SNR using 3D lifting, detailed in the following, reveals that it and hence the derived ROI selection~\cite{Lempe.2013, Kim.2021} is strongly influenced by the specific illumination setup. This dependency underscores the critical importance of considering lighting conditions when selecting ROIs for accurate pulse measurement.
A few observations can be made from the resulting 3D pulse maps and the diffuse skin map in FLAME texture space (see \Cref{fig:texture_summary}): The SNR is high in skin regions, while regions covered with facial hair and eyes show low SNR. The green channel amplitude is mostly uniform over the face, with the exception of the eyes. 
At the nasal bone, where skin thickness exhibits a minimum, both SNR and amplitude are decreased. 
The phase (both green channel and estimated using POS) is uniform over most of the facial skin. A phase shift can be observed at the lips and the eye regions, and a phase inversion (phase$\pm 180°$) can be observed at the neck in the vicinity of the carotid artery.
This phase inversion can be attributed to a different physical signal being measured~\cite{Moco.2017}: While in the face slight color changes of the skin due to blood perfusion are the source of the pulse signal (the classical PPGI signal), on the neck the source is the mechanical micro-motion of the skin induced by the dominant contribution of the carotid artery wall.

We validate our approach qualitatively by assessing changes in the pulse maps, and quantitatively by evaluating the FLAME reconstruction, the illumination scenario (``diffuse vs ring light''), and the resulting texture maps given all influences (directional illumination, accuracy of camera alignment, accuracy of FLAME fitting, accuracy of texture building).

\paragraph{Validity of pulse maps:}
To validate the pulse maps, we applied a slight scratch to the skin of subject S02 by gently rubbing the finger nail over the center of the forehead (see ~\Cref{fig:ex_validity}a). The scratch is visible in the RGB image as a reddish mark. All maps show changes in the region of the scratch, with the most extreme changes in the SNR map.

The 3D-projected SNR maps reveal features that are not discernible in a single 2D SNR map. For instance, a blemish that was barely visible in the RGB domain becomes clearly distinguishable on the right cheek in the texture representation shown in \Cref{fig:ex_validity}b.

\begin{figure}
    \centering
    \includegraphics[trim={0.5cm 1cm 0 1cm},clip,width=0.8\linewidth]{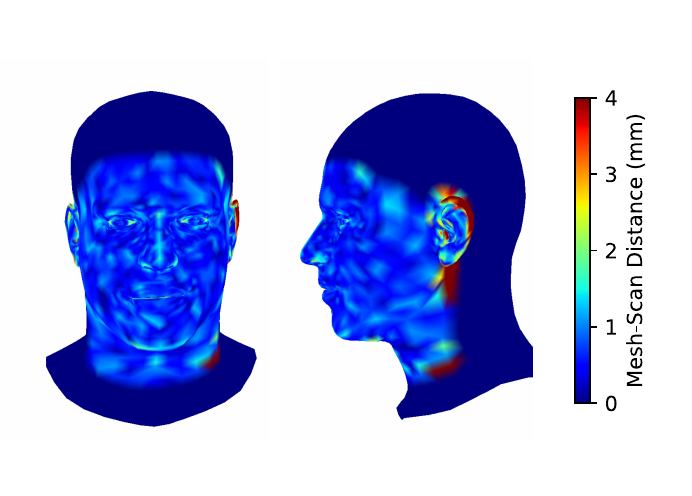}
    \caption{Mesh to scan error in face and neck region (subject S01).}
    \label{fig:mesh_error}
\end{figure}

\begin{figure}[h!]
    \centering
    \includegraphics[width=\linewidth]{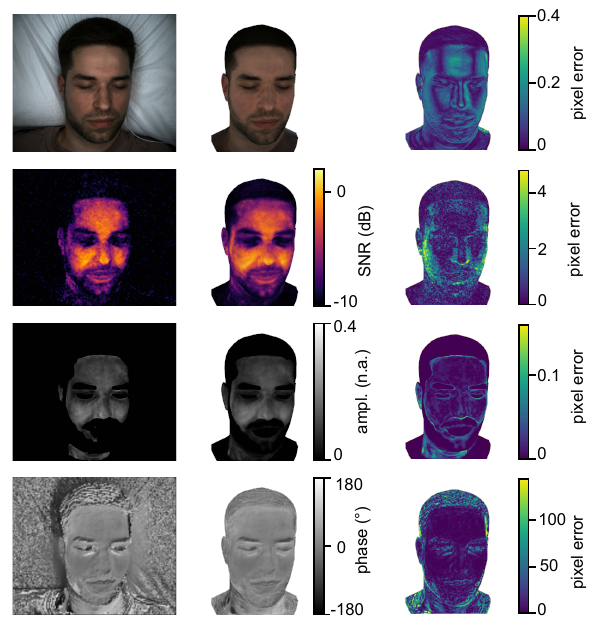}      
    \caption{Visualization of reprojection error of base maps.}
    \label{fig:reprojection_error_vis}
\end{figure}

\paragraph{FLAME fitting and texture validity:}
The error of the fitted FLAME mesh stays below 1\,mm in most of the facial skin region (see~\Cref{fig:mesh_error}).  Larger errors exist at the facial boundaries, the ears, and the neck where our measurement setup provides no or low quality data.
Facial boundaries and edges of facial features such as the nose are susceptible to motion and thus typically not reliable for PPGI.
Assuming the physiological consistency of the maps from multiple views, we propose to compare the reprojection error of the respective SNR, amplitude, and phase maps with that of the diffuse RGB texture for all 23 views (\Cref{fig:reprojection_error_vis}, top-down view).
In RGB, the main error sources are the mesh error and specular highlights that are removed by the averaging. The SNR shows a similar dependency on illumination. Amplitude and phase are less dependent on illumination and more consistent over the face overall, therefore show a generally smaller error. In case of the phase, we can observe that the phase inversion at the neck is reduced by the averaging, thus resulting in a higher error.

\begin{figure}[h!]
    \centering
    \includegraphics[width=0.9\linewidth]{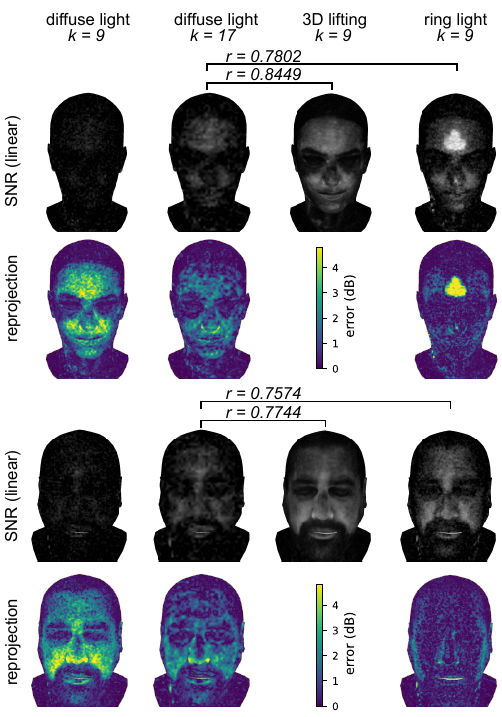}
    \caption{Evaluation of illumination setting; comparing diffuse to ring light with different $k \times k$ sliding windows. We compute Pearson correlation $r$ for SNR to compare the similarity between the (optimal) diffuse light 2D map, the ring light 2D map, and the 3D lifted map from the ring light.}
    \label{fig:reprojection_error_vis_light}
\end{figure}

\paragraph{Influence of illumination:}
In an ideal scenario, the illumination would be perfectly diffuse, and the resulting pulse maps would accurately represent the true pulsatility of the skin.
In our (and most other) real set-ups, the closest approximation to diffuse illumination is achieved by using large softboxes.
While the resulting pulse maps appear homogeneous, they suffer from rather low SNR due to the difficulty of simultaneously providing high light power and uniform illumination. 
This low SNR can only be enhanced by increasing the window size $k$ for pulse extraction, which in turn reduces the spatial resolution. 
The ring light on the other hand provides strong illumination but introduces local highlights in the pulse maps.
\Cref{fig:reprojection_error_vis_light} presents the frontal view of $M_\text{SNR}$ under diffuse illumination for window size $k=9$ and for $k=17$. A usable SNR distribution is observed for $k=17$. This is compared to the 3D lifting computed from the ring light condition (here $k=9$) and the corresponding 2D-map. To highlight differences, the reprojection error between the 3D lifting and each 2D-map is shown.
Based on the Pearson correlation, the 3D lifting achieves a closer approximation of the diffuse illumination condition than directly using the 2D-map under ring-light illumination. 
The highlights induced by non-uniform light are effectively reduced by our texture-averaging approach for obtaining 3D-maps.
The effect is most pronounced in SNR and RGB maps.
For the phase and amplitude maps, the error of diffuse and ring light are in a similar range, and both are not affected by highlights due to the normalization.

\section{Discussion and Limitations}

\subsection{Measurement Setup} 
\textbf{Illumination}: The ring light provides high SNR in facial regions closer to the light source for most subjects, enabling the generation of high-resolution pulse maps. 
However, the intensity drop-off results in uneven illumination, which affects amplitude and phase estimation despite normalization by light intensity. 
An optimal setup would involve high-power, diffuse lighting covering the semi-spherical recording area to ensure uniform illumination.
One fundamental limitation of this work (and any other work of this type) is that there is no independent ground truth measurement system to obtain pulsation maps. However, by using 3D lifting, we obtain SNR maps that average out the influence of lighting effects. At the same time, local perturbations known to increase (scratch) or decrease (pimple) perfusion are preserved and more distinguishable in 3D. 

\noindent\textbf{Camera System}: The robot arm with a single sensor offered a flexible solution for capturing multiple views instead of requiring a dedicated multi-camera setup. 
However, for certain subjects (e.g., S04 and S11), excessive head and facial motion led to poor camera alignment, difficulties in FLAME fitting due to discrepancies in mouth and jaw positions across views, and ultimately, misaligned textures.
These issues could be mitigated by employing a more sophisticated multi-camera system, as demonstrated by~\cite{Kirschstein.2023}.
Polarization filters have been used in~\cite{Azinovic.2023} and others to compute specular reflection maps on top of albedo, normal maps, and facial geometry. As algorithms such as POS use specular reflection to enhance robustness to motion, we did not consider such filters. Still, it might be worth evaluating if they might improve on the single color channel analysis.

\subsection{Reconstruction}
The FLAME model was selected for its sufficient resolution in mapping accurate pulse data onto facial skin regions. 
Subjects were instructed to keep their eyelids closed to minimize motion artifacts from blinking. 
However, FLAME does not accurately model eyelid positions, which could be improved by incorporating an eye closure loss function, as proposed by~\cite{Feng.2021}.
Heavy subject motion, such as non-rigid motion of nostrils, mouth and neck, that degrade registration, could be addressed by leveraging morphable models generated from single-view analysis using an analysis-by-synthesis approach~\cite{Retsinas.2024}.
For the application of phase maps in facial perfusion simulation, the current texture fusion method, which averages data across multiple views, may introduce inaccuracies. 
Specifically, this averaging process smooths phase inversions, biasing angles closer to 0°, potentially reducing the fidelity of the reconstructed maps. 
Depending on the use case, some of the maps such as amplitude or phase maps require postprocessing. Especially near edges and head boundaries, phase shifts or strong amplitudes can be observed that are thought to be caused by ballistocardiographic effects~\cite{Zaunseder.2018}.
These are one reason, why we applied skin masks to the visualization of the amplitude maps.
These limitations should be considered when applying the maps in further research.


\section{Conclusion}
In this work, we introduced \ours, to our knowledge, the first dataset for estimating 3D blood pulsation maps from RGB video. Furhtermore, we proposed 3D blood puslation maps and presented a prototypical pipeline for computing SNR, pulse amplitude, and phase across all three color channels, alongside skin segmentation masks in the FLAME texture space. 
Our evaluation demonstrates that these maps are physiologically plausible and that mapping pulsation data into 3D space reveals information that was previously difficult to assess in 2D.
We also made a first effort in quantitatively evaluating the resulting 3D maps, which is a difficult task due to the lack of a well defined ground truth.
The dataset, including texture maps and raw data, is publicly available.

\section*{Acknowledgments} 
We are grateful to Jacob Otto for helping with the development and operation of the measurement setup and to the many participants for allowing their facial data to be shared and contribute to our research.
The authors acknowledge the Lichtenberg high performance computing cluster of the TU Darmstadt for providing computational facilities for the calculations of this research. 

{
    \small
    \bibliographystyle{ieeenat_fullname}
    \bibliography{main}

@inproceedings{Alexander.2009,
 author = {Alexander, Oleg and Rogers, Mike and Lambeth, William and Chiang, Matt and Debevec, Paul},
 title = {The Digital Emily project},
 pages = {12:1--12:15},
 publisher = {ACM},
 isbn = {9781450379380},
 booktitle = {ACM SIGGRAPH 2009 Courses},
 year = {2009},
 address = {New York, NY, USA},
 doi = {10.1145/1667239.1667251}
}

@inproceedings{Moco.2018,
 author = {Mo{\c{c}}o, Andreia Vieira and Stuijk, Sander and {van Gastel}, Mark and de Haan, Gerard},
 title = {Impairing Factors in Remote-PPG Pulse Transit Time Measurements on the Face},
 editor = {IEEE},
 booktitle = {IEEE Conference on Computer Vision and Pattern Recognition Workshops},
 year = {2018}
}

@article{Moco.2017,
 author = {Mo{\c{c}}o, Andreia Vieira and Zavala-Mondragon, Luis Albert and Wang, Wenjin and Stuijk, Sander and de Haan, Gerhard},
 year = {2017},
 title = {Camera-based assessment of arterial stiffness and wave reflection parameters from neck micro-motion},
 pages = {1576--1598},
 volume = {38},
 number = {8},
 journal = {Physiological Measurement},
 doi = {10.1088/1361-6579/aa7d43}
}

@inproceedings{Nogue.2024,
 author = {Nogue, Emilie and lin, Arvin and Li, Xiaouhui and Guarnera, Giuseppe Claudio and Ghosh, Abhijeet},
 title = {Practical RGB Measurement of Fluorescence and Blood Distributions in Skin},
 publisher = {IS{\&}T},
 booktitle = {Color and Imaging Conference},
 year = {2024}
}

@inproceedings{Nowara.2018,
 author = {Nowara, Ewa Magdalena and Marks, Tim K. and Mansour, Hassan and Veeraraghavan, Ashok},
 title = {SparsePPG: Towards Driver Monitoring Using Camera-Based Vital Signs Estimation in Near-Infrared},
 pages = {1353--135309},
 publisher = {IEEE},
 isbn = {978-1-5386-6100-0},
 booktitle = {CVPRW},
 year = {2018},
 doi = {10.1109/CVPRW.2018.00174}
}

@inproceedings{Qian.2024,
 author = {Qian, Shenhan and Kirschstein, Tobias and Schoneveld, Liam and Davoli, Davide and Giebenhain, Simon and Nie{\ss}ner, Matthias},
 title = {GaussianAvatars: Photorealistic Head Avatars with Rigged 3D Gaussians},
 pages = {20299--20309},
 publisher = {IEEE},
 booktitle = {2024 IEEE/CVF Conference on Computer Vision and Pattern Recognition (CVPR)},
 year = {2024},
 doi = {10.1109/cvpr52733.2024.01919}
}

@inproceedings{Retsinas.2024,
 author = {Retsinas, George and Filntisis, Panagiotis P. and Dan{\v{e}}{\v{c}}ek, Radek and Abrevaya, Victoria F. and Roussos, Anastasios and Bolkarr, Timo and Maragos, Petros},
 title = {3D Facial Expressions through Analysis-by-Neural-Synthesis},
 pages = {2490--2501},
 publisher = {IEEE},
 booktitle = {2024 IEEE/CVF Conference on Computer Vision and Pattern Recognition (CVPR)},
 year = {2024},
 doi = {10.1109/cvpr52733.2024.00241}
}

@inproceedings{Rohr.2024,
 author = {Rohr, Maurice and Witulla, Philipp and {Hoog Antink}, Christoph},
 title = {Skin Reflection Angle Useful for Region of Interest Selection in Camera-based Heart Rate Estimation?},
 editor = {IEEE},
 booktitle = {Computing in Cardiology},
 year = {2024}
}

@article{Sabour.2023,
 author = {Sabour, Rita Meziati and Benezeth, Yannick and de Oliveira, Pierre and Chapp{\'e}, Julien and Yang, Fan},
 year = {2023},
 title = {UBFC-Phys: A Multimodal Database For Psychophysiological Studies of Social Stress},
 pages = {622--636},
 volume = {14},
 number = {1},
 issn = {1949-3045},
 journal = {IEEE Transactions on Affective Computing},
 doi = {10.1109/TAFFC.2021.3056960}
}

@article{Scherpf.2024,
 author = {Scherpf, Matthieu and Ernst, Hannes and Malberg, Hagen and Schmidt, Martin},
 year = {2024},
 title = {DeepPerfusion: A Comprehensible Two-Branched Deep Learning Architecture for High-Precision Blood Volume Pulse Extraction Based on Imaging Photoplethysmography},
 journal = {techrxiv},
 doi = {10.36227/techrxiv.172503790.08194939/v1}
}

@article{Schraven.2023,
 author = {Schraven, Sebastian P. and Kossack, Benjamin and Str{\"u}der, Daniel and Jung, Maximillian and Skopnik, Lotte and Gross, Justus and Hilsmann, Anna and Eisert, Peter and Mlynski, Robert and Wisotzky, Eric L.},
 year = {2023},
 title = {Continuous intraoperative perfusion monitoring of free microvascular anastomosed fasciocutaneous flaps using remote photoplethysmography},
 pages = {1532},
 volume = {13},
 number = {1},
 issn = {2045-2322},
 journal = {Scientific reports},
 doi = {10.1038/s41598-023-28277-w}
}

@article{Tan.2023,
 author = {Tan, Chengyifeng and Xiao, Chang and Wang, Wenjin},
 year = {2023},
 title = {Camera-based Cardiovascular Screening based on Heart Rate and Its Variability In Pre- and Post-Exercise Conditions},
 pages = {1--5},
 volume = {2023},
 journal = {Annual International Conference of the IEEE Engineering in Medicine and Biology Society. IEEE Engineering in Medicine and Biology Society. Annual International Conference},
 doi = {10.1109/EMBC40787.2023.10340871}
}

@article{Teplov.2014,
 author = {Teplov, Victor and Nippolainen, Ervin and Makarenko, Alexander A. and Giniatullin, Rashid and Kamshilin, Alexei A.},
 year = {2014},
 title = {Ambiguity of mapping the relative phase of blood pulsations},
 pages = {3123--3139},
 volume = {5},
 number = {9},
 issn = {2156-7085},
 journal = {Biomedical optics express},
 doi = {10.1364/BOE.5.003123}
}

@article{Wang.2017,
 author = {Wang, Wenjin and den Brinker, Albertus C. and Stuijk, Sander and de Haan, Gerard},
 year = {2017},
 title = {Algorithmic Principles of Remote PPG},
 pages = {1479--1491},
 volume = {64},
 number = {7},
 issn = {1558-2531},
 journal = {IEEE Transactions on Biomedical Engineering},
 doi = {10.1109/TBME.2016.2609282}
}

@article{Wang.2024,
 author = {Wang, Haowen and Huang, Jia and Wang, Guowei and Lu, Hongzhou and Wang, Wenjin},
 year = {2024},
 title = {Contactless Patient Care Using Hospital IoT: CCTV-Camera-Based Physiological Monitoring in ICU},
 pages = {5781--5797},
 volume = {11},
 number = {4},
 journal = {IEEE Internet of Things Journal},
 doi = {10.1109/JIOT.2023.3308477}
}

@inproceedings{Yang.2015,
 author = {Yang, Jun and Guthier, Benjamin and {El Saddik}, Abdulmotaleb},
 title = {Estimating two-dimensional blood flow velocities from videos},
 pages = {3768--3772},
 publisher = {IEEE},
 isbn = {978-1-4799-8339-1},
 booktitle = {2015 IEEE International Conference on Image Processing (ICIP)},
 year = {2015},
 doi = {10.1109/ICIP.2015.7351509}
}

@article{Yang.2024,
 author = {Yang, Lingchen and Zoss, Gaspard and Chandran, Prashanth and Gross, Markus and Solenthaler, Barbara and Sifakis, Eftychios and Bradley, Derek},
 year = {2024},
 title = {Learning a Generalized Physical Face Model From Data},
 pages = {1--14},
 volume = {43},
 number = {4},
 issn = {0730-0301},
 journal = {ACM Transactions on Graphics},
 doi = {10.1145/3658189}
}

@inproceedings{Zaunseder.2018,
author = {Sebastian Zaunseder and Alexander Trumpp and Hannes Ernst and Michael F{\"o}rster and Hagen Malberg},
title = {{Spatio-temporal analysis of blood perfusion by imaging photoplethysmography}},
volume = {10501},
booktitle = {Optical Diagnostics and Sensing XVIII: Toward Point-of-Care Diagnostics},
editor = {Gerard L. Cot{\'e}},
organization = {International Society for Optics and Photonics},
publisher = {SPIE},
pages = {105010X},
year = {2018},
doi = {10.1117/12.2289896}
}

@inproceedings{Zhang.2023,
 author = {Zhang, Tianke and Chu, Xuangeng and Liu, Yunfei and Lin, Lijian and Yang, Zhendong and Xu, Zhengzhuo and Cao, Chengkun and Yu, Fei and Zhou, Changyin and Yuan, Chun and Li, Yu},
 title = {Accurate 3D Face Reconstruction with Facial Component Tokens},
 publisher = {IEEE},
 booktitle = {2023 IEEE/CVF International Conference on Computer Vision (ICCV)},
 year = {2023},
 doi = {10.1109/iccv51070.2023.00829}
}

@article{Shao.2014,
 author = {Shao, Dangdang and Yang, Yuting and Liu, Chenbin and Tsow, Francis and Yu, Hui and Tao, Nongjian},
 year = {2014},
 title = {Noncontact monitoring breathing pattern, exhalation flow rate and pulse transit time},
 pages = {2760--2767},
 volume = {61},
 number = {11},
 issn = {1558-2531},
 journal = {IEEE Transactions on Biomedical Engineering},
 doi = {10.1109/TBME.2014.2327024}
}

@incollection{McDuff.2022,
 author = {McDuff, Daniel and Wander, Miah and Liu, Xin and Hill, Brian L. and Hernandez, Javier and Lester, Jonathan and Baltrusaitis, Tadas},
 title = {SCAMPS: Synthetics for Camera Measurement of Physiological Signals},
 publisher = {{Curran Associates, Inc.}},
 booktitle = {NeurIPS},
 year = {2022},
 doi = {10.48550/arXiv.2206.04197}
}

@article{McDuff.2023,
 author = {McDuff, Daniel},
 year = {2023},
 title = {Camera Measurement of Physiological Vital Signs},
 pages = {1--40},
 volume = {55},
 number = {9},
 issn = {0360-0300},
 journal = {ACM Computing Surveys},
 doi = {10.1145/3558518}
}

@inproceedings{Azinovic.2023,
 author = {Azinovi{\'c}, Dejan and Maury, Olivier and Hery, Christophe and Nie{\ss}ner, Matthias and Thies, Justus},
 title = {High-Res Facial Appearance Capture from Polarized Smartphone Images},
 pages = {16836--16846},
 publisher = {IEEE},
 isbn = {979-8-3503-0129-8},
 booktitle = {2023 IEEE/CVF Conference on Computer Vision and Pattern Recognition (CVPR)},
 year = {2023},
 doi = {10.1109/CVPR52729.2023.01615}
}

@inproceedings{Ba.2022,
 author = {Ba, Yunhao and Wang, Zhen and Karinca, Kerim Doruk and Bozkurt, Oyku Deniz and Kadambi, Achuta},
 title = {Style Transfer with Bio-realistic Appearance Manipulation for Skin-tone Inclusive rPPG},
 pages = {1--12},
 publisher = {IEEE},
 isbn = {978-1-6654-5851-1},
 booktitle = {2022 IEEE International Conference on Computational Photography (ICCP)},
 year = {2022},
 doi = {10.1109/ICCP54855.2022.9887649}
}

@article{Bazarevsky.2019,
 author = {Bazarevsky, Valentin and Kartynnik, Yury and Vakunov, Andrey and Raveendran, Karthik and Grundmann, Matthias},
 year = {2019},
 title = {BlazeFace: Sub-millisecond Neural Face Detection on Mobile GPUs},
 journal = {arXiv},
 doi = {10.48550/arXiv.1907.05047}
}

@inproceedings{Blanz.1999,
 author = {Blanz, Volker and Vetter, Thomas},
 title = {A morphable model for the synthesis of 3D faces},
 pages = {187--194},
 publisher = {{ACM Press}},
 booktitle = {Proceedings of the 26th annual conference on Computer graphics and interactive techniques  - SIGGRAPH '99},
 year = {1999},
 address = {New York, New York, USA},
 doi = {10.1145/311535.311556}
}

@article{Feng.2021,
 author = {Feng, Yao and Feng, Haiwen and Black, Michael J. and Bolkart, Timo},
 year = {2021},
 title = {Learning an animatable detailed 3D face model from in-the-wild images},
 volume = {40},
 number = {4},
 journal = {ACM Transactions on Graphics},
 doi = {10.1145/3450626.3459936}
}

@incollection{Frassineti.2017,
 author = {Frassineti, Lorenzo and Giardini, Francesco and Perrella, Antonia and Sorelli, Michele and Sacconi, Leonardo and Bocchi, Leonardo},
 title = {Evaluation of spatial distribution of skin blood flow using optical imaging},
 pages = {74--80},
 volume = {62},
 publisher = {{Springer Singapore}},
 isbn = {978-981-10-4165-5},
 series = {IFMBE Proceedings},
 editor = {Badnjevic, Almir},
 booktitle = {CMBEBIH 2017},
 year = {2017},
 address = {Singapore},
 doi = {10.1007/978-981-10-4166-2_12}
}

@article{Gotardo.2018,
 author = {Gotardo, Paulo and Riviere, J{\'e}r{\'e}my and Bradley, Derek and Ghosh, Abhijeet and Beeler, Thabo},
 year = {2018},
 title = {Practical dynamic facial appearance modeling and acquisition},
 pages = {1--13},
 volume = {37},
 number = {6},
 issn = {0730-0301},
 journal = {ACM Transactions on Graphics},
 doi = {10.1145/3272127.3275073}
}

@article{Haan.2013,
 author = {de Haan, Gerard and Jeanne, Vincent},
 year = {2013},
 title = {Robust pulse rate from chrominance-based rPPG},
 pages = {2878--2886},
 volume = {60},
 number = {10},
 journal = {IEEE transactions on bio-medical engineering},
 doi = {10.1109/TBME.2013.2266196}
}

@article{Hammer.2022,
 author = {Hammer, Alexander and Scherpf, Matthieu and Schmidt, Martin and Ernst, Hannes and Malberg, Hagen and Matschke, Klaus and Dragu, Adrian and Martin, Judy and Bota, Olimpiu},
 year = {2022},
 title = {Camera-based assessment of cutaneous perfusion strength in a clinical setting},
 volume = {43},
 number = {2},
 journal = {Physiological Measurement},
 doi = {10.1088/1361-6579/ac557d}
}

@article{Haugg.2022,
 author = {Haugg, Fridolin and Elgendi, Mohamed and Menon, Carlo},
 year = {2022},
 title = {Effectiveness of Remote PPG Construction Methods: A Preliminary Analysis},
 volume = {9},
 number = {10},
 issn = {2306-5354},
 journal = {Bioengineering (Basel, Switzerland)},
 doi = {10.3390/bioengineering9100485}
}

@article{Jeong.2016,
 author = {Jeong, In Cheol and Finkelstein, Joseph},
 year = {2016},
 title = {Introducing Contactless Blood Pressure Assessment Using a High Speed Video Camera},
 pages = {77},
 volume = {40},
 number = {4},
 journal = {Journal of medical systems},
 doi = {10.1007/s10916-016-0439-z}
}

@inproceedings{Jimenez.2010,
 author = {Jimenez, Jorge and Weyrich, Tim and Scully, Timothy and Barbosa, Nuno and Donner, Craig and Alvarez, Xenxo and Vieira, Teresa and Matts, Paul and Orvalho, Ver{\'o}nica and Gutierrez, Diego},
 title = {A practical appearance model for dynamic facial color},
 pages = {1},
 publisher = {{ACM Press}},
 isbn = {9781450304399},
 editor = {Drettakis, George},
 booktitle = {ACM SIGGRAPH Asia 2010 papers on - SIGGRAPH ASIA '10},
 year = {2010},
 address = {New York, New York, USA},
 doi = {10.1145/1866158.1866167}
}

@article{Kamshilin.2011,
 author = {Kamshilin, Alexei A. and Miridonov, Serguei and Teplov, Victor and Saarenheimo, Riku and Nippolainen, Ervin},
 year = {2011},
 title = {Photoplethysmographic imaging of high spatial resolution},
 pages = {996--1006},
 volume = {2},
 number = {4},
 issn = {2156-7085},
 journal = {Biomedical optics express},
 doi = {10.1364/BOE.2.000996}
}

@article{Kim.2021,
 author = {Kim, Dae-Yeol and Lee, Kwangkee and Sohn, Chae-Bong},
 year = {2021},
 title = {Assessment of ROI Selection for Facial Video-Based rPPG},
 pages = {7923},
 volume = {21},
 number = {23},
 journal = {Sensors (Basel, Switzerland)},
 doi = {10.3390/s21237923}
}

@article{Kirschstein.2023,
 author = {Kirschstein, Tobias and Qian, Shenhan and Giebenhain, Simon and Walter, Tim and Nie{\ss}ner, Matthias},
 year = {2023},
 title = {NeRSemble: Multi-view Radiance Field Reconstruction of Human Heads},
 pages = {1--14},
 volume = {42},
 number = {4},
 issn = {0730-0301},
 journal = {ACM Transactions on Graphics},
 doi = {10.1145/3592455}
}

@incollection{Kossack.2022,
 author = {Kossack, Benjamin and Wisotzky, Eric and Eisert, Peter and Schraven, Sebastian P. and Globke, Brigitta and Hilsmann, Anna},
 title = {Perfusion assessment via local remote photoplethysmography (rPPG)},
 pages = {2191--2200},
 volume = {64},
 publisher = {IEEE},
 booktitle = {2022 IEEE/CVF Conference on Computer Vision and Pattern Recogntion Workshops},
 year = {2022},
 doi = {10.1109/CVPRW56347.2022.00238}
}

@article{Lai.2022,
 author = {Lai, Marco and {van der Stel}, Stefan D. and Groen, Harald C. and {van Gastel}, Mark and Kuhlmann, Koert F. D. and Ruers, Theo J. M. and Hendriks, Benno H. W.},
 year = {2022},
 title = {Imaging PPG for In Vivo Human Tissue Perfusion Assessment during Surgery},
 volume = {8},
 number = {4},
 journal = {Journal of imaging},
 doi = {10.3390/jimaging8040094}
}

@incollection{Lempe.2013,
 author = {Lempe, Georg and Zaunseder, Sebastian and Withgen, Tom and Zipse, Stephan and Malberg, Hagen},
 title = {ROI Selection for Remote Photoplethymography},
 booktitle = {Bildverarbeitung f{\"u}r die Medizin},
 publisher = {Springer},
 year = {2013}
}

@inproceedings{Li.2018,
 author = {Li, Xiaobai and Alikhani, Iman and Shi, Jingang and Seppanen, Tapio and Junttila, Juhani and Majamaa-Voltti, Kirsi and Tulppo, Mikko and Zhao, Guoying},
 title = {The OBF Database: A Large Face Video Database for Remote Physiological Signal Measurement and Atrial Fibrillation Detection},
 pages = {242--249},
 publisher = {IEEE},
 isbn = {978-1-5386-2335-0},
 booktitle = {2018 13th IEEE International Conference on Automatic Face {\&} Gesture Recognition},
 year = {2018},
 doi = {10.1109/FG.2018.00043}
}

@article{Li.2017,
 author = {Li, Tianye and Bolkart, Timo and Black, Michael J. and Li, Hao and Romero, Javier},
 year = {2017},
 title = {Learning a model of facial shape and expression from 4D scans},
 pages = {1--17},
 volume = {36},
 number = {6},
 issn = {0730-0301},
 journal = {ACM Transactions on Graphics},
 doi = {10.1145/3130800.3130813}
}

@inproceedings{Zhang.2016,
 author = {Zhang, Zheng and Girard, Jeffrey M. and Wu, Yue and Zhang, Xing and Liu, Peng and Ciftci, Umur and Canavan, Shaun and Reale, Michael and Horowitz, Andrew and Yang, Huiyuan and Cohn, Jeffrey F. and Ji, Qiang and Yin, Lijun},
 title = {Multimodal Spontaneous Emotion Corpus for Human Behavior Analysis},
 publisher = {IEEE},
 booktitle = {2016 IEEE Conference on Computer Vision and Pattern Recognition (CVPR)},
 year = {2016},
 doi = {10.1109/cvpr.2016.374}
}

@article{Zollhofer.2018,
 author = {Zollh{\"o}fer, M. and Thies, J. and Garrido, P. and Bradley, D. and Beeler, T. and P{\'e}rez, P. and Stamminger, M. and Nie{\ss}ner, M. and Theobalt, C.},
 year = {2018},
 title = {State of the Art on Monocular 3D Face Reconstruction, Tracking, and Applications},
 pages = {523--550},
 volume = {37},
 number = {2},
 issn = {0167-7055},
 journal = {Computer Graphics Forum},
 doi = {10.1111/cgf.13382}
}

@inproceedings{Scherpf.2021,
 author = {Scherpf, Matthieu and Ernst, Hannes and Misera, Leo and Malberg, Hagen and Schmidt, Martin},
 title = {Skin Segmentation for Imaging Photoplethysmography Using a Specialized Deep Learning Approach},
 booktitle = {Computing in Cardiology},
 year = {2021}
}

@article{Zielonka.2022,
 author = {Zielonka, Wojciech and Bolkart, Timo and Thies, Justus},
 year = {2022},
 title = {Towards Metrical Reconstruction of Human Faces},
 journal = {European Conference on Computer Vision}
}

@article{Pstras.2025,
 author = {Pstras, Leszek and Okupnik, Tymoteusz and Ponikowska, Beata and Paleczny, Bartlomiej},
 year = {2025},
 title = {Facial video photoplethysmography for measuring average and instantaneous heart rate: a pilot validation study},
 journal = {medRxiv},
 doi = {10.1101/2025.02.13.25322005}
}

@article{Bonnet.2022,
 author = {Bonnet, S. and Lubin, M. and Doron, M. and Blanquer, G. and Perriollat, M. and Prada, R. and Blandin, P. and Gerbelot, R.},
 year = {2022},
 title = {Spatial dependency of the PPG morphology at right carotid common artery},
 pages = {3146--3149},
 volume = {2022},
 journal = {Annual International Conference of the IEEE Engineering in Medicine and Biology Society. IEEE Engineering in Medicine and Biology Society. Annual International Conference},
 doi = {10.1109/EMBC48229.2022.9871985}
}

@incollection{Loper.2014,
 author = {Loper, Matthew M. and Black, Michael J.},
 title = {OpenDR: An Approximate Differentiable Renderer},
 pages = {154--169},
 volume = {8695},
 publisher = {{Springer International Publishing}},
 isbn = {978-3-319-10583-3},
 series = {Lecture Notes in Computer Science},
 editor = {Fleet, David and Pajdla, Tomas and Schiele, Bernt and Tuytelaars, Tinne},
 booktitle = {Computer Vision -- ECCV 2014},
 year = {2014},
 address = {Cham},
 doi = {10.1007/978-3-319-10584-0_11}
}

@book{Nocedal.2006,
 author = {Nocedal, Jorge and Wright, Stephen J.},
 year = {2006},
 title = {Numerical Optimization},
 publisher = {{Springer New York}},
 isbn = {978-0-387-30303-1},
 doi = {10.1007/978-0-387-40065-5}
}

@inproceedings{Wu.2000,
 author = {Wu, Ting and Blazek, Vladimir and Schmitt, Hans Juergen},
 title = {Photoplethysmography imaging: a new noninvasive and noncontact method for mapping of the dermal perfusion changes},
 pages = {62--70},
 publisher = {SPIE},
 booktitle = {Optical Techniques and Instrumentation for the Measurement of Blood Composition, Structure, and Dynamics},
 year = {2000},
 doi = {10.1117/12.407646}
}

@article{Wang.2025,
 author = {Wang, Haowen and Huang, Jia and Wang, Guowei and Lou, Jincan and Tancheng, Yifeng and Lu, Dongmin Huang Hongzhou and Wang, Wenjin},
 year = {2025},
 title = {Towards Camera-PRV Based Early Warning in Hospital ICU: A Pilot Study},
 pages = {1},
 journal = {IEEE Internet of Things Journal},
 doi = {10.1109/JIOT.2025.3566990}
}

@incollection{Braun.2024,
 author = {Braun, Bj{\"o}rn and McDuff, Daniel and Holz, Christian},
 title = {How Suboptimal is Training rPPG Models with Videos and Targets from Different Body Sites?},
 booktitle = {Proceedings of the IEEE/CVF Conference on Computer Vision and Pattern Recognition (CVPR) Workshops},
 year = {2024},
 doi = {10.48550/arXiv.2403.10582}
}
}
\clearpage
\setcounter{page}{1}
\maketitlesupplementary

\begin{figure*}
    \centering
    \includegraphics[width=\linewidth]{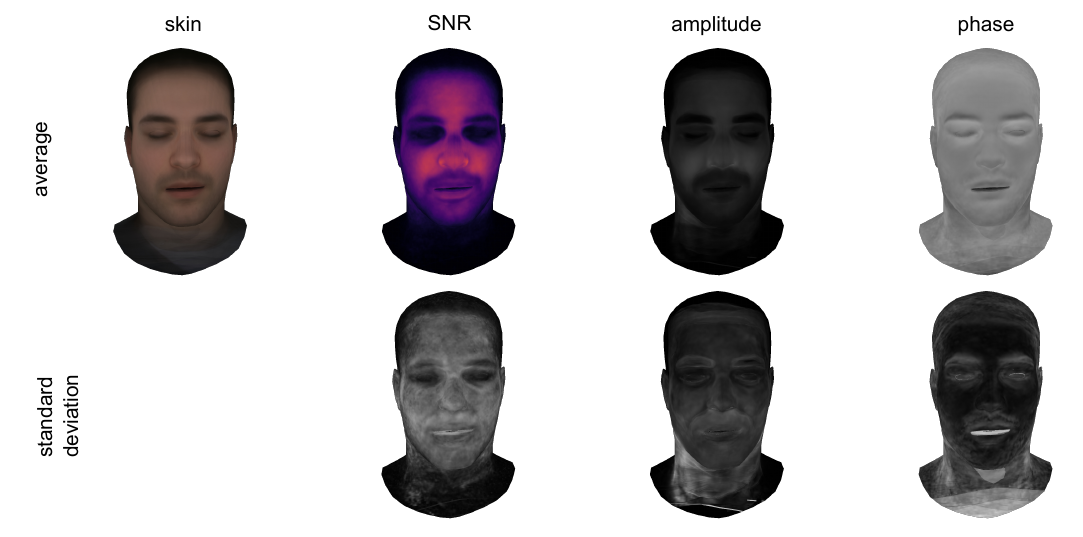}
    \caption{Rendered maps averaged over all 15 subjects and their standard deviations showing key areas that are similar for all subjects, such as lower SNR on nasal bone and bearded regions. Sharp features such as phase inversion are not visible. Eye region and mouth show a different phase than the rest of the facial skin. Notice the larger standard deviation in regions typically covered by facial hair in men. The phase is computed from the green channel ($M_\text{a,g}$)}
    \label{fig:supp_average_maps}
\end{figure*}%

\section{Additional Results}
\subsection{Pulse Map Statistical Analysis}
We compute mean and standard deviation of SNR, amplitude, and phase textures to show the regions that are similar between different subjects, and regions where the largest differences occur (Figure~\ref{fig:supp_average_maps}). As described in the paper, decreased SNR and amplitude can be observed on average in the nasal bone region. The largest differences between subjects can be observed in the beard and eyebrow regions.

\subsection{Pulse Map Views}

\begin{figure*}
    \centering
    \includegraphics[width=\linewidth]{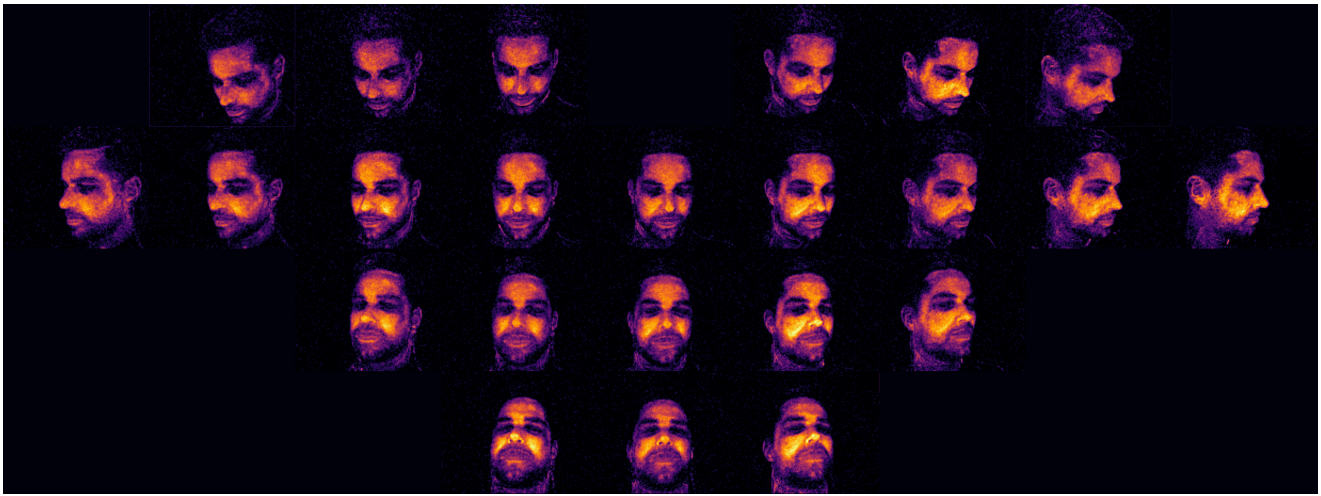}
    \caption{SNR maps from all 23 viewing directions. More vibrant color means higher SNR. Notice how the SNR is depending on view and illumination. High SNR spots are mostly close to the center of the camera where the distance from face to the ring light is minimal. Surface parts approximately orthogonal to the camera achieve very low SNR as do parts covered by significant facial hair.}
    \label{fig:snr_views}
\end{figure*}

\begin{figure*}
    \centering
    \includegraphics[width=\linewidth]{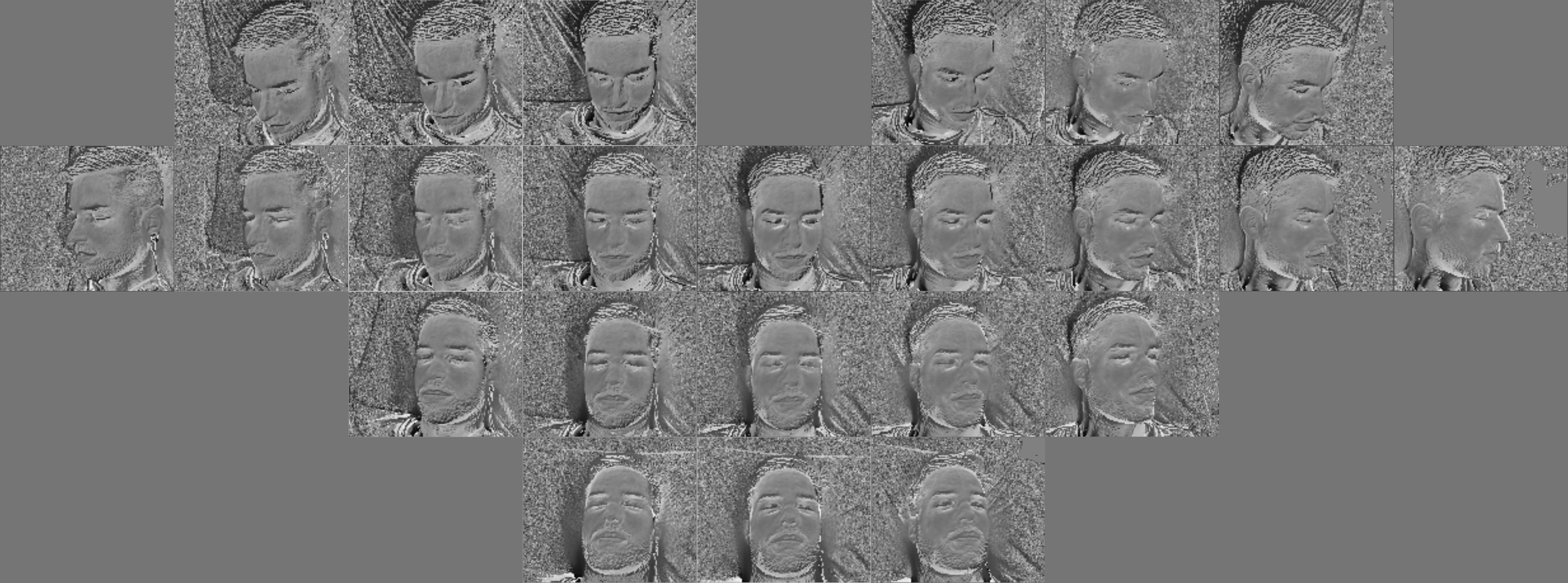}
    \caption{Phase maps of green color channel $M_\text{p,g}$ from all 23 viewing directions scaled from $-\pi,\pi$. Grey represents zero phase. Notice the phase inversion on the neck caused by the different source of the signal. The signal on the facial skin is mostly due to slight color variations of the skin caused by pulsatile blood perfusion, while the signal in the vicinity of the carotid artery at the neck is caused by illumination variation due to mechanical changes of the skin surface due to displacement of the arterial walls. }
    \label{fig:phase_views}
\end{figure*}

\begin{figure*}
    \centering
    \includegraphics[width=\linewidth]{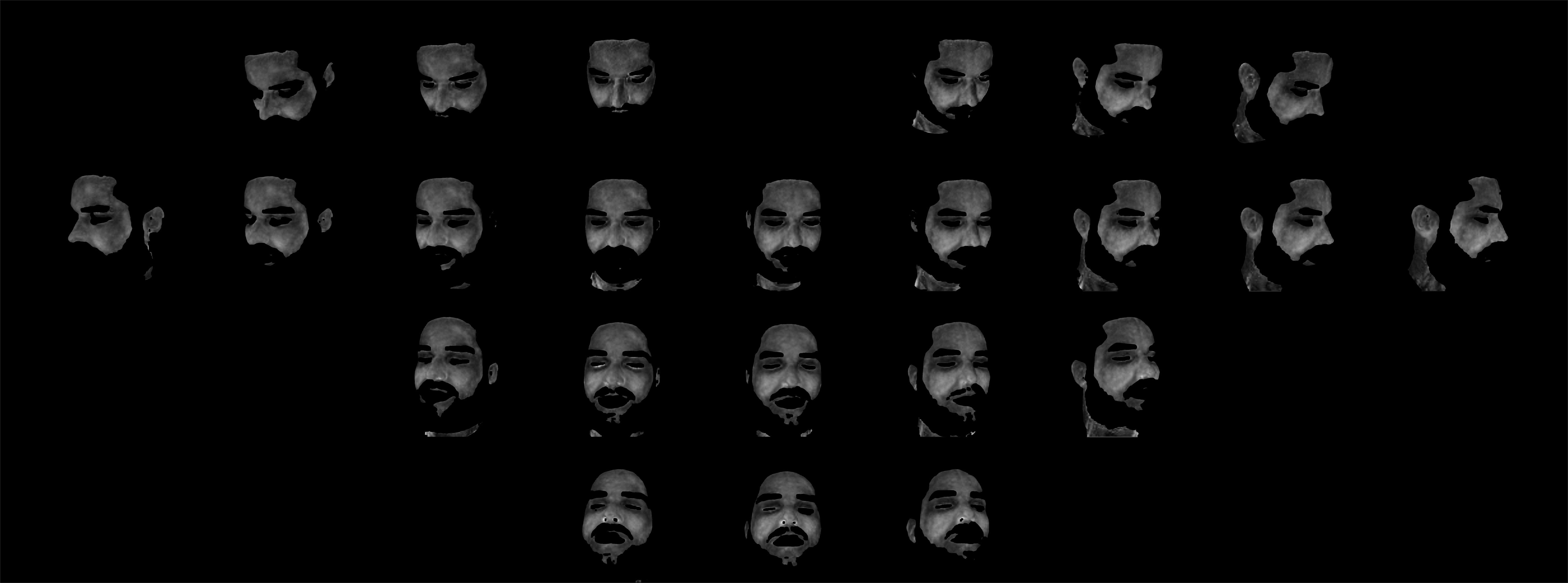}
    \caption{Amplitude maps of green color channel $M_\text{a,g}$ from all 23 viewing directions. Brighter color means higher amplitude. We applied the skin masks to highlight changes across the skin. Along the borders of the face, as well as hairlines and other features, the amplitude is often increased due to motion artifacts. Also, ballistocardiographic effects caused by the acceleration of the head in the direction of the cardiac ejection can cause strong amplitudes along edges. Notice that the amplitude is less dependent on illumination than the SNR depicted in Figure~\ref{fig:ampl_views}.}
    \label{fig:ampl_views}
\end{figure*}

The pulse maps from all recorded 23 views for subject S01 are depicted in Figures~\ref{fig:snr_views},\ref{fig:phase_views} and \ref{fig:ampl_views}.

\subsection{Additional Pulse Maps}

\begin{figure*}
    \centering
    \includegraphics[width=\linewidth]{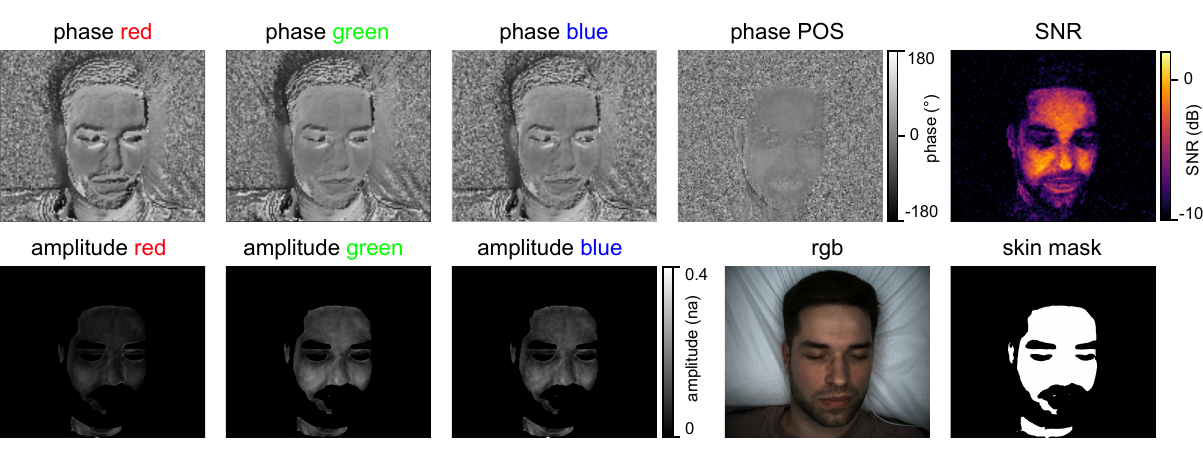}
    \caption{All computed maps depicted for subject S01 from the central view: phase maps $M_{\text{p},c}$ and amplitude maps $M_{\text{a},c}$ of all color channels $c$, phase map compute from POS signal via FFT, SNR map, first rgb-frame of video and skin mask. Brighter colors represent higher values. We applied the skin masks also to the amplitude maps to highlight changes across the skin. Notice the differences between the amplitude maps. The green color channel shows the highest signal amplitude, blue's amplitude is slightly lower and the result is more noisy. Red shows the lowest amplitude. The phase maps are very similar between the color channels. In the \emph{phase POS} map, the phase difference between lips and skin is most prominent. Such phase differences are expected if tissue is connected to different parts of the arterial tree~\cite{Teplov.2014}.}
    \label{fig:all_maps}
\end{figure*}

For the main results, we selected representative maps. The entire selection of maps that we computed is depicted in Figure~\ref{fig:all_maps}

\subsection{Texture Maps}

\begin{figure*}
    \centering
    \includegraphics[width=\linewidth]{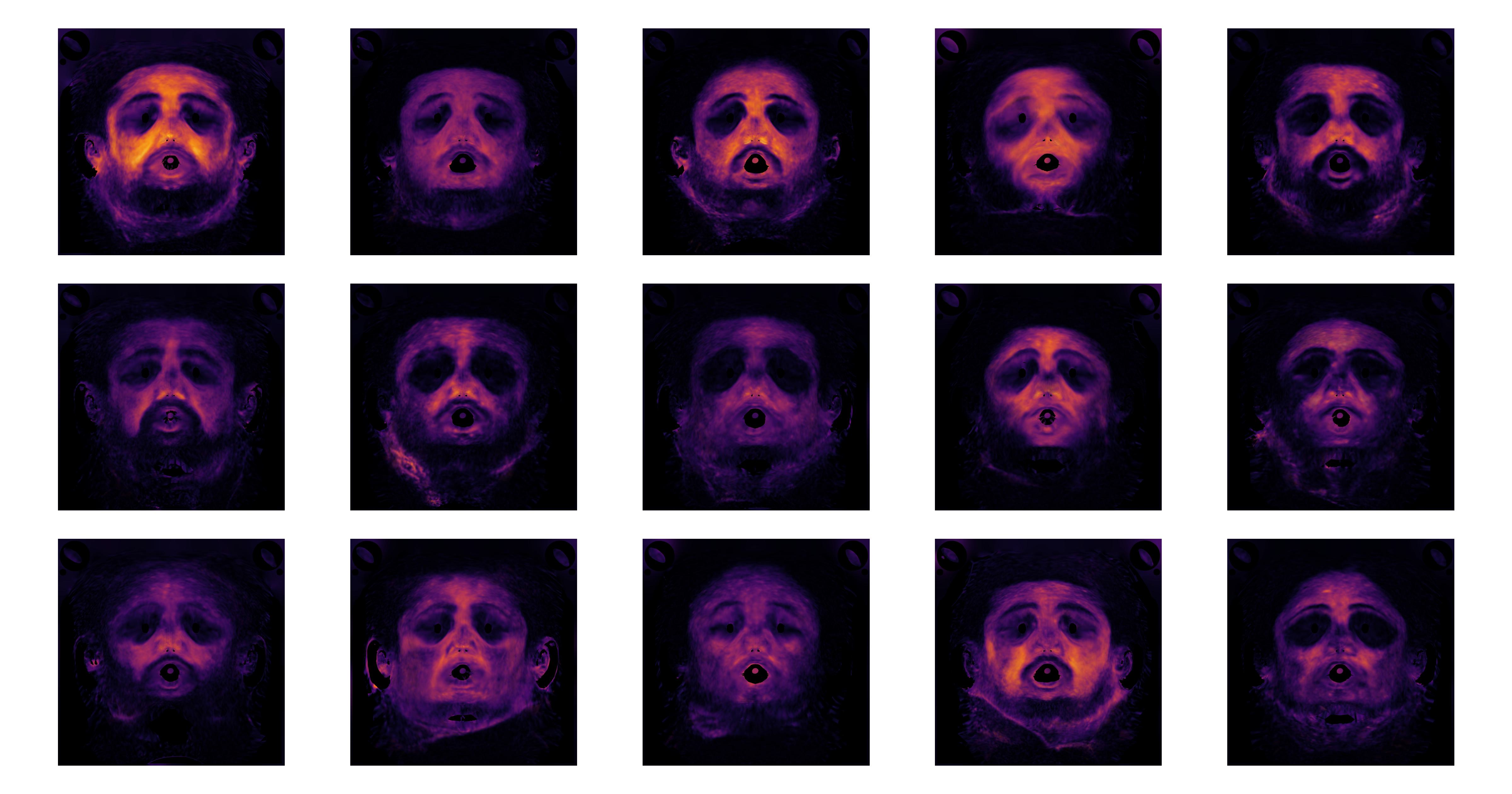}
    \caption{SNR textures for FLAME  for all 15 subjects. More vibrant color means higher SNR. }
    \label{fig:snr_textures}
\end{figure*}

\begin{figure*}
    \centering
    \includegraphics[width=\linewidth]{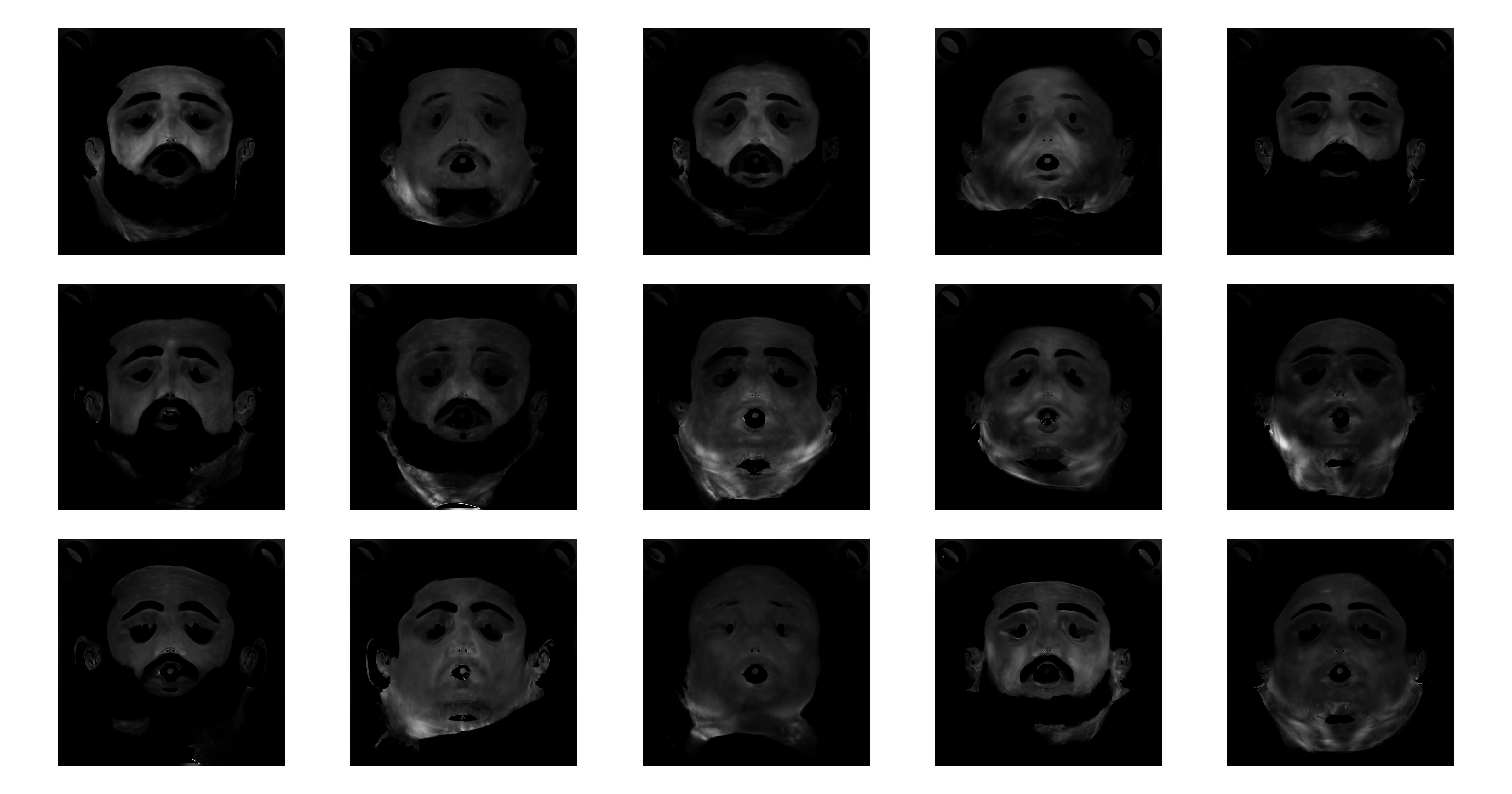}
    \caption{Green channel amplitude textures for FLAME  for all 15 subjects. }
    \label{fig:ampl_textures}
\end{figure*}

\begin{figure*}
    \centering
    \includegraphics[width=\linewidth]{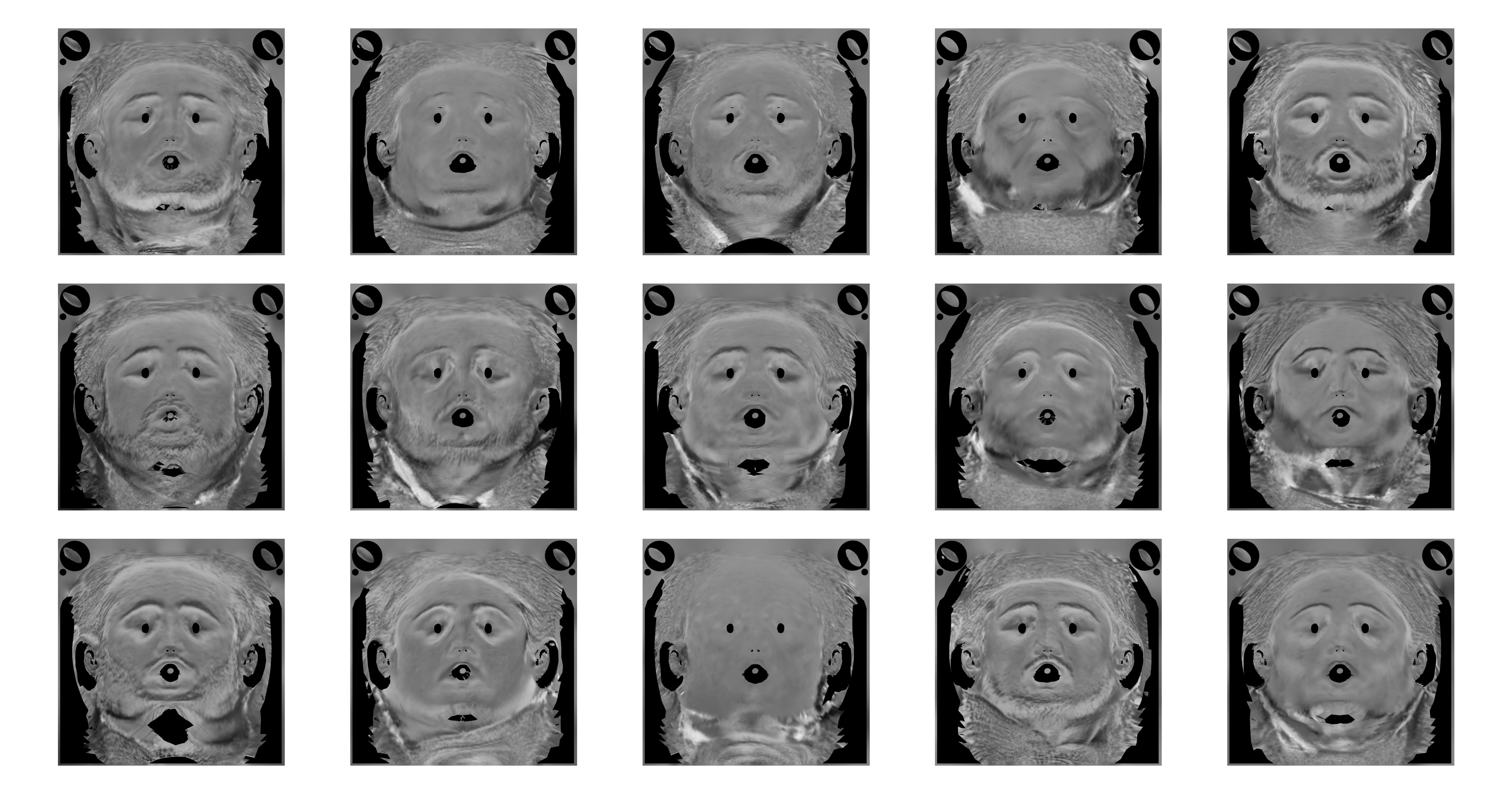}
    \caption{Green channel phase textures for FLAME  for all 15 subjects.}
    \label{fig:phase_textures}
\end{figure*}

We show example texture maps for all 15 subjects consisting of SNR~(Figure~\ref{fig:snr_textures}), amplitude~(Figure~\ref{fig:ampl_textures})  and phase~(Figure~\ref{fig:phase_textures}). Notice how pulse amplitude vary quite strongly between subject. Also notice that the maximum SNR per subject varies quite strongly even though the recording conditions were controlled, with identical illumination and camera settings across all subjects. The phase textures show hints of phase inversion at the neck, when the neck is not covered. When the neck is covered, the texture is noise in that region.

\section{Additional Evaluation}

\begin{figure*}[htp]
\centering
\includegraphics[width=.32\textwidth]{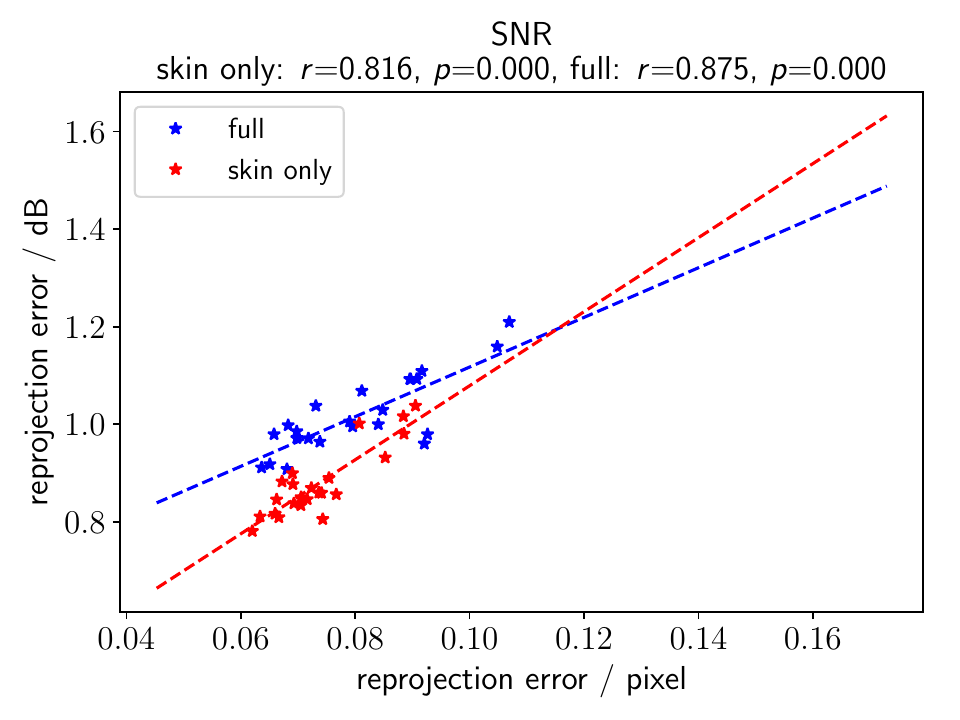}\hfill
\includegraphics[width=.32\textwidth]{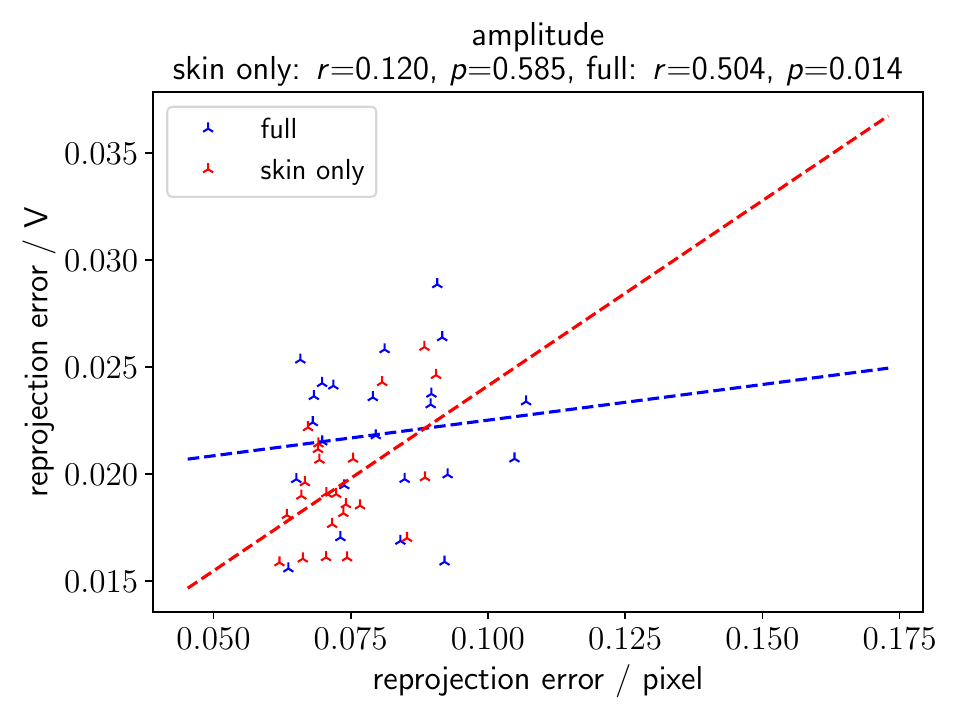}\hfill
\includegraphics[width=.32\textwidth]{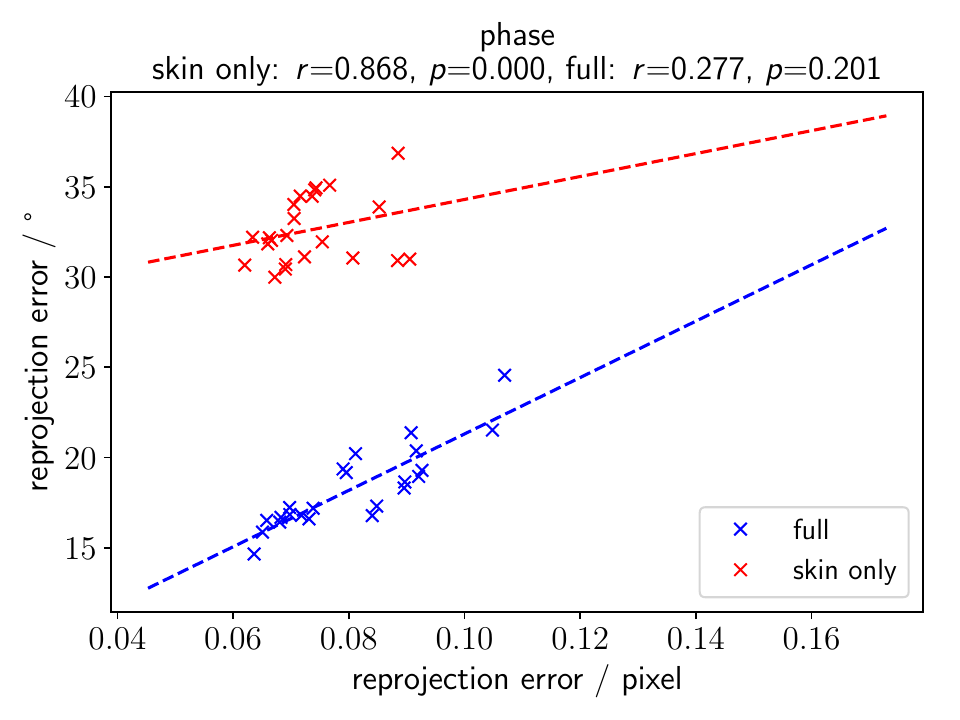}
\caption{ Average root-mean-square reprojection error per view for the SNR, amplitude ($M_\text{a,g}$), and phase ($M_\text{p,g}$) maps in correlation to the  reprojection error of the diffuse texture (computed from RGB images). All plots show the reprojection error computed based on the full FLAME surface and only the skin regions based on our computed skin masks. We found a significant correlation ($p<0.05$ and $|r|>0.2$) between the SNR/phase reprojection error and the diffuse reprojection error for the skin region. The amplitude reprojection error shows no significant correlation with the diffuse reprojection error for the skin region. Note that we excluded S04 and S11 since their recordings show strong motion artifacts.}
\label{fig:view_repro_error}
\end{figure*}

\subsection{Comparison with Previous Work}

\begin{figure*}
    \centering
    \includegraphics[width=\linewidth]{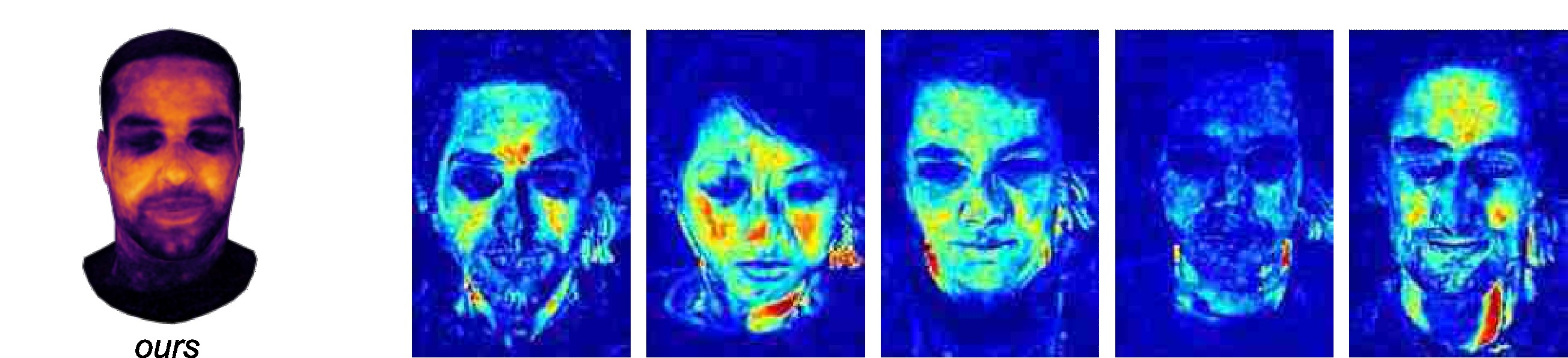}
    \caption{Pulse map computed by \cite{Lempe.2013} in comparison to \emph{ours}. Notice that the SNR is heavily influenced by highlights caused by uneven illumination and quantitative results might not be suitable to find the physiological optimal ROI independent of illumination.}
    \label{fig:snr_lempe}
\end{figure*}

We compare our results using 3D lifting and our specialized dataset with a representative example from previous studies~\cite{Lempe.2013} (Figure~\ref{fig:snr_lempe}). The comparison of the SNR underscores the critical importance of considering lighting conditions when selecting ROIs for accurate pulse measurement.

\subsection{Goodness of Fit}
\begin{figure*}
    \centering
    \includegraphics[width=\linewidth]{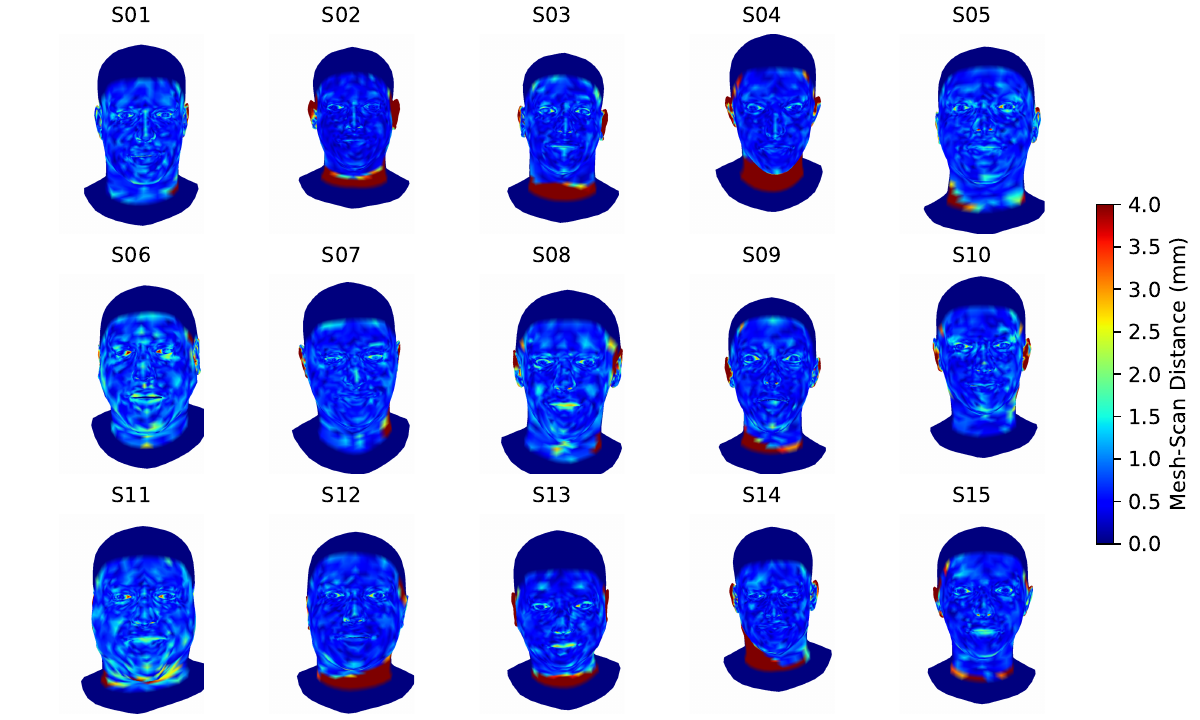}
    \caption{Mesh-to-scan errors computed for all 15 subjects. The facial skin region is most important for pulse mapping and shows errors below 1\,mm almost everywhere. Hair, clothing, and for some subjects the ears, were excluded from FLAME fitting.}
    \label{fig:goodness_fit}
\end{figure*}

An essential assumption of our work is that we can model all faces with a generic face model with a shared texture space. The accuracy of the final texture maps depends on the camera alignment, low facial movement, and the ability to fit the FLAME model to the SFM scan mesh.
For completeness, we compute the mesh-to-scan error for all 15 subjects~(Figure~\ref{fig:goodness_fit}). The FLAME mesh shows sub-millimeter accuracy for most facial skin regions. Larger errors exist at the facial boundaries, the ears, and the neck where our measurement setup provides no or low quality data. For some subjects, such as S12, the neck was covered by clothing, making fitting the model in these regions unfeasible. For subject S11, we noticed extensive motion during the recording, which influences camera alignment, pulse map estimation, and model fitting as shown here. 

\subsection{Dependency Analysis of Reprojection Error}

We are interested in estimating accurate pulse maps. However, there is no independent ground truth to compare our results to. Additionally, there are multiple factors that can deteriorate the quality of the estimated 3D pulse maps: Non-uniform illumination affects the 2D pulse map estimates, camera alignment, and the mesh-to-scan error affect the 3D estimation. We assume illumination effects to be view-dependent, whereas camera alignment and mesh-to-scan error are affected mainly by motion, which we assume to be subject-dependent. For example, in case of the phase, we can observe that the phase inversion at the neck is reduced by our current implementation through averaging, thus resulting in a higher error.
Because we are only interested in error specific to the pulse maps, we try to mitigate these effects by comparing the reprojection error calculated for the pulse maps based on the 3D pulse map and the 2D pulse map from known views, with the reprojection error of the regular diffuse skin texture (using the same pipeline for texture estimation).
We correlate the reprojection error of SNR, amplitude ($M_\text{a,g}$), and phase ($M_\text{p,g}$) maps to the reprojection error of the diffuse texture for all 23 views (Figure~\ref{fig:view_repro_error}) using averaged root-mean-square values. The plots show both the reprojection errors computed on the full model (``full'') and on the skin region computed with our skin segmentation only (``skin only'').  We regard dependency as relevant if $p<0.05$ and $|r|>0.2$.
The correlation between reprojection error of diffuse texture and the SNR texture is high, meaning that there are certain angles that generally lead to higher errors, for both representations that we regarded as strongly influenced by illumination. The dependency analysis of the reprojection error of diffuse texture and the amplitude/phase textures is inconclusive. We can make out that the amplitude is generally less view dependent as the error does not increase with the diffuse reprojection error if only the skin regions are considered. Remember that the amplitude is normalized by illumination intensity, and thus less sensitive to illumination than the SNR for example.
However, more analysis is required to come to a definitive conclusion.


\end{document}